\documentclass[10pt,twocolumn,letterpaper]{article}

\usepackage[final]{iccv}      %

\usepackage{float}
\usepackage[utf8]{inputenc} %
\usepackage[T1]{fontenc}    %
\usepackage{url}            %
\usepackage{booktabs}       %
\usepackage{amsfonts}       %
\usepackage{nicefrac}       %
\usepackage{microtype}      %
\usepackage{bbm}
\usepackage[dvipsnames,table]{xcolor}
\usepackage[framemethod=tikz]{mdframed}

\usepackage{multirow}       %
\usepackage{amsmath,amssymb} %

\usepackage{symbols}

\usepackage{siunitx}  %
\usepackage{amsmath,amsfonts,bm}
\usepackage{amsthm,amssymb}

\def\1{\bm{1}}

\def\sp{space}

\def\rvz{{\mathbf{z}}}

\def\rmF{{\mathbf{F}}}

\def\rmR{{\mathbf{R}}}
\def\rmS{{\mathbf{S}}}
\def\rmT{{\mathbf{T}}}

\def\vzero{{\bm{0}}}

\def\mO{{\bm{O}}}

\def\mS{{\bm{S}}}

\DeclareMathAlphabet{\mathsfit}{\encodingdefault}{\sfdefault}{m}{sl}
\SetMathAlphabet{\mathsfit}{bold}{\encodingdefault}{\sfdefault}{bx}{n}

\def\gD{{\mathcal{D}}}

\def\gG{{\mathcal{G}}}
\def\gH{{\mathcal{H}}}

\def\gL{{\mathcal{L}}}
\def\gM{{\mathcal{M}}}

\def\gS{{\mathcal{S}}}

\def\gU{{\mathcal{U}}}

\def\gW{{\mathcal{W}}}
\def\gX{{\mathcal{X}}}
\def\gY{{\mathcal{Y}}}

\def\sR{{\mathbb{R}}}

\newcommand{\E}{\mathbb{E}}

\newcommand*{\ShowNotes}{} %
\definecolor{darkred}{rgb}{0.7,0.1,0.1}
\definecolor{darkgreen}{rgb}{0.1,0.7,0.1}
\definecolor{cyan}{rgb}{0.7,0.0,0.7}
\definecolor{dblue}{rgb}{0.2,0.2,0.8}
\definecolor{maroon}{rgb}{0.76,.13,.28}
\definecolor{burntorange}{rgb}{0.81,.33,0}
\definecolor{tealblue}{rgb}{0.212,0.459, 0.533}

\definecolor{mypink}{rgb}{0.93359375, 0.62109375, 0.83984375}

\definecolor{pp}{rgb}{0.43921569, 0.18823529, 0.62745098}
\definecolor{rr}{rgb}{0.5254902 , 0.00784314, 0.12941176}
\definecolor{bb}{rgb}{0.09019608, 0.23529412, 0.37647059}
\definecolor{yy}{rgb}{0.49803922, 0.3372549 , 0.0}
\definecolor{gg}{rgb}{0.02352941, 0.3372549 , 0.17647059}
\definecolor{aliceblue}{rgb}{0.94, 0.97, 1.0}
\definecolor{OursColor}{rgb}{0.89,0.96, 1.}

\ifdefined\ShowNotes
  \newcommand{\colornote}[3]{{\color{#1}\bf{#2: #3}\normalfont}}
\else
  \newcommand{\colornote}[3]{}
\fi

\newcommand {\baselineA}{{\it Aug}}
\newcommand {\baselineB}{{\it Canon}}
\newcommand {\baselineC}{{\it InvL}}
\newcommand {\baselineD}{{\it AugL}}
\newcommand{\generic}{{\it Base}}

\newcommand{\ourrow}{\rowcolor{OursColor}}

\newcommand{\myparagraph}[1]{\vspace*{2pt}{\bf\noindent #1.}}

\usepackage[many]{tcolorbox}

\tcolorboxenvironment{definition}{
  boxrule=0pt,
  boxsep=0pt,
  colback={White!90!LimeGreen},
  enhanced jigsaw, 
  borderline west={1pt}{0pt}{LimeGreen},
  sharp corners,
  before skip=10pt,
  after skip=10pt,
  left=5pt,
  right=5pt,
  breakable,
}

\tcolorboxenvironment{myexp}{
  boxrule=0pt,
  boxsep=0pt,
  colback={White!90!White},
  enhanced jigsaw, 
  borderline west={1pt}{0pt}{BurntOrange},
  sharp corners,
  before skip=3pt,
  after skip=6pt,
  left=5pt,
  right=5pt,
  bottom=1pt,
  top=3pt,
  breakable,
}

\tcolorboxenvironment{myclaim}{
  boxrule=0pt,
  boxsep=0pt,
  colback={White!90!White},
  enhanced jigsaw, 
  borderline west={1pt}{0pt}{White},
  sharp corners,
  before skip=10pt,
  after skip=10pt,
  left=5pt,
  right=5pt,
  breakable,
}

\newtcolorbox{myclaime}[1][]{
  notitle, %
  #1, %
  boxrule=0pt,
  boxsep=0pt,
  colback={White!90!White},
  enhanced jigsaw, 
  borderline west={1pt}{0pt}{CornflowerBlue},
  sharp corners,
  before skip=2pt,
  after skip=3pt,
  left=5pt,
  right=5pt,
  bottom=1pt,
  top=3pt,
  breakable,
}

\tcolorboxenvironment{example}{
  boxrule=0pt,
  boxsep=0pt,
  blanker,
  borderline west={1pt}{0pt}{Black},
  sharp corners,
  before skip=10pt,
  after skip=10pt,
  left=5pt,
  right=5pt,
  breakable,
}

\tcolorboxenvironment{mylemma}{
  boxrule=0pt,
  boxsep=0pt,
  colback={White!90!Gray},
  enhanced jigsaw, 
  borderline west={1pt}{0pt}{Gray},
  sharp corners,
  before skip=10pt,
  after skip=10pt,
  left=5pt,
  right=5pt,
  breakable,
}

\newtcolorbox{mylemmae}[1][]{
  notitle, %
  #1, %
  boxrule=0pt,
  boxsep=0pt,
  colback={White!96!Blue},
  enhanced jigsaw, 
  borderline west={1.5pt}{0pt}{Gray},
  sharp corners,
  before skip=1pt,
  after skip=3pt,
  left=5pt,
  right=5pt,
  bottom=1pt,
  top=2pt,
  breakable,
}

\usepackage[multiple]{footmisc}

\usepackage{thm-restate}

\definecolor{iccvblue}{rgb}{0.21,0.49,0.74}
\usepackage[pagebackref,breaklinks,colorlinks,allcolors=iccvblue]{hyperref}

\def\paperID{9054} %
\def\confName{ICCV}
\def\confYear{2025}

\title{Local Scale Equivariance with Latent Deep Equilibrium Canonicalizer}

\author{
    Md Ashiqur Rahman$^1$ \quad
    Chiao-An Yang$^1$ \quad
    Michael N. Cheng$^1$ \quad
    Lim Jun Hao$^2$ \\
    Jeremiah Jiang$^2$  \quad
    Teck-Yian Lim$^2$ \quad
    Raymond A. Yeh$^1$
    \\ 
    $^1$Department of Computer Science, Purdue University \quad $^2$DSO National Laboratories
}

\begin{document}
\twocolumn[{%
\renewcommand\twocolumn[1][]{#1}%
\maketitle
}]

\begin{abstract}
Scale variation is a fundamental challenge in computer vision. Objects of the same class can have different sizes, and their perceived size is further affected by the distance from the camera. These variations are local to the objects, \ie, different object sizes may change differently within the same image. To effectively handle scale variations, we present a deep equilibrium canonicalizer (DEC) to improve the local scale equivariance of a model. DEC can be easily incorporated into existing network architectures and can be adapted to a pre-trained model. Notably, we show that on the competitive ImageNet benchmark, DEC improves both model performance and local scale consistency across four popular pre-trained deep-nets, \eg, ViT, DeiT, Swin, and BEiT. 
Our code is available at \url{https://github.com/ashiq24/local-scale-equivariance}.
\end{abstract}

\section{Introduction}

The perceived scale of an object in images varies due to multiple factors. First, intrinsic factors such as the object's physical dimensions cause variations in the apparent size,~\eg, individuals may differ in height. Second, extrinsic factors such as camera zoom and the object's distance to the camera further affect the observed scale, even for the same object.
How to effectively handle and represent this scale variation has been a fundamental question in computer vision. Inspired by human vision, which uses a hierarchy of scales for perception~\cite{marr2010vision}, earlier works adopted image pyramids~\cite{adelson1984pyramid,burt1987laplacian} and variants~\cite{537667,witkin1987signal,lindeberg2013scale} to extract features at different image resolutions~\cite{lowe1999object,lowe2004distinctive,grauman_2005_pyramid,lazebnik2006beyond}.

In deep learning, ideas similar to image pyramids have been introduced~\cite{he2015spatial,zhao2017pyramid, chen2017deeplab,fan2021multiscale} and studied under the broad area of scale invariance/equivariance~\cite{sosnovik2019scale,worrall2019deep,sosnovik2021disco,rahman2024truly}. These works aim to design specialized deep-nets such that the network's output is transformed in a predefined way when the input image undergoes scale changes (image resizing). Mathematically, equivariance is formalized as equality statements over groups; see~\secref{sec:prelim} for a review. A typical approach is to design a group equivariant architecture~\cite{cohen2016group} by modifying the filter operations. Note, we use the term ``equivariance'' in the strict mathematical sense of the \textit{exact equality statement}. We will use the term ``consistency'' to mean an approximation of equivariance,~\ie,  without the mathematical guarantees. 

While scale equivariance has been well studied, these works all focused on {\it global scaling}, \ie, changes in image resolution. Differently, we are interested in {\it local scaling}, \eg, the resizing of each part of an input image differently (at a fixed resolution), which more closely matches the scaling of objects in the real world. See illustration in~\figref{fig:teaser}. %
\begin{figure}[t]
    \input{figs/teaser}
\end{figure}
Ideally, we aim to design deep-nets that are local scale equivariant. However, real-world local scaling does not form a group as the operation is not invertible,~\eg, when two objects change size and occlude each other. To address this, we propose %
an approximation of the real-world local scaling operation, namely monotone scaling, which we will show to be a group. 

Our approach to making a deep-net monotone scale equivariant is based on the framework of canonicalization~\cite{mondal2023equivariant, panigrahi2024improved}. Briefly, a canonicalizer transforms an input into a ``standard'' (canonical) form so that the deep network can extract invariant features. %
A canonicalizer can be modeled as finding a stationary point of a learned energy function with respect to the transformation (group) and then using that point as the ``standard''. Instead of this optimization approach, we propose to utilize a deep equilibrium model~\cite{bai2019deep} for directly predicting this point,~\ie, amortized optimization~\cite{amos2023tutorial}. We name our approach the Deep Equilibrium Canonicalizer (DEC). We then incorporate DEC into the latent space of existing deep-net architectures to improve the local scale consistency.

To evaluate local scale consistency, we first conduct experiments on semantic segmentation using synthetic data where we can control the scale variation. We create a dataset of realistic renderings using physical simulation based on the Google Scan Objects~\cite{downs2022google}. Next, we consider the task of image classification. Inspired by the MNIST-scale dataset used in prior works~\cite{kanazawa2014locally,sosnovik2021disco,rahman2024truly}, we create a dataset of locally scaled MNIST images where each image contains multiple MNIST digits, resized to a different scale. Finally, we evaluate a locally scaled version of ImageNet~\cite{deng2009imagenet}. 
Overall, our proposed DEC, when incorporated into existing models, achieves a higher performance metric and further improves the local scale consistency over the existing base model.

{\bf \noindent Our main contributions are as follows:}
\begin{itemize}
    \item Motivated by local scaling in the real world, we propose to study how to design deep-nets that are equivariant to monotone scaling.
    \item We propose DEC, a novel latent canonicalization approach based on deep equilibrium models, which can be easily incorporated into existing deep-net architectures.
    \item We demonstrate that the proposed DEC improves model performance and local scale consistency on two vision tasks, notably, the competitive benchmark of ImageNet across four different ViT architectures.
\end{itemize}

\section{Related Works}

{\bf\noindent Scale consistency} has a long history in computer vision. Early works focus on developing methods that are invariant to scaling~\cite{lowe2004distinctive, witkin1987scale} with techniques such as Laplacian image pyramids~\cite{1095851, 537667} or wavelet analysis~\cite{van2008marr}. Later works aim to improve scale consistency by learning transformations applied either to the input images or directly to convolutional filters~\cite{jaderberg2015spatial, recasens2018learning, thavamani2021fovea, dai2017deformable} or by directly learning local/global scale estimation \cite{lee2022self, barroso2022scalenet}. More recently, variants of vision transformer models have been proposed to tackle scale consistency, \eg, incorporating the structure of image pyramids~\cite{gu2022multi, tian2023resformer}. While some of these works use the term invariance/equivariance, these approaches generally do not achieve the equality condition on a group,~\ie, they are not guaranteed to be scale equivariant in the strict sense.  In the equivariant literature, there is a line of work studying global scaling~\cite{xu2014scale,worrall2019deep,sosnovik2019scale,sosnovik2021disco,rahman2024truly} through designing specialized architecture based on group theory.

\myparagraph{Group equivariant models}
Designing group equivariant architectures has been explored due to their strong theoretical guarantees and robustness~\cite{cohen2016group, cesa2022program, puny2021frame, ma2024canonicalization, yeh2022equivariance}. Their development has been extended to incorporate equivariance in rotation~\cite{weiler2018learning}, shift~\cite{rojas2024making, zhang2019making},
permutation~\cite{ravanbakhsh_sets,zaheer2017deep,hartford2018deep,yeh2019chirality,liu2021semantic,
liu2020pic,yeh2019diverse},
and sampling~\cite{xu2021group,rojas2022learnable, rahman2025group}. Global scale equivariant architectures are primarily constructed by integrating over the scale space of images~\cite{worrall2019deep}, through the use of steerable filter~\cite{sosnovik2019scale} or applying spectral-domain operations for anti-aliasing \cite{rahman2024truly}. However, the current literature lacks equivariant methods for local scaling (as it does not form a group) and only handles the global scaling operations.

\myparagraph{Equivariant via canonicalization} With the rise of pre-trained models, there is growing interest in adapting existing models to be equivariant rather than developing entirely new equivariant architectures. This has led to the area of group equivariant adaptation, which primarily aims to ``standardize'' the input through the use of a canonicalization module \cite{mondal2023equivariant, panigrahi2024improved} or by averaging over the group elements to achieve more stable output \cite{basu2024efficient, basu2023equi}. However, these methods target discrete symmetry groups with a few parameters,~\eg, the $2D$ rotation group ($SO(2)$) characterized by a single parameter. These approaches are limited when the size and dimensionality of the transformation group increases. %

{\bf \noindent Deep Equilibrium Models (DEQs)}
 are a class of implicit neural networks that compute their outputs by solving for the fixed point of a parameterized transformation~\cite{bai2019deep, bai2020multiscale}. Unlike traditional deep networks with a predefined number of layers, DEQs effectively model an infinitely deep, weight-tied recurrent network, making them highly parameter-efficient. These models have been applied successfully in various applications, including optical flow estimation~\cite{bai2022deep} and video landmark detection~\cite{micaelli2023recurrence}. In this work, we found DEQs to be a suitable architecture for modeling a canonicalizer for adapting deep-nets to be equivariant.

\section{Preliminaries}\label{sec:prelim}
 We provide a review of the definition of equivariance and invariance. For readability, we describe these concepts using one-dimensional functions. Next, we review the canonicalization framework for achieving equivariant models.

\myparagraph{Global scaling operation}
The global scaling operation $\rmR_a$ on a 1D function $f \in \gU$, where $\gU \triangleq \{f: [0,1] \rightarrow \sR\}$, is defined as 
\bea
\label{eqn:glob_scaling}
\rmR_a[f](x) = f(a^{-1}x),
\eea
where $a \in \sR^+$ is the scaling factor. Specifically, $\rmR_a$ with $a>1$ is an upscaling operation, and $a<1$ is a downscaling operation.

\myparagraph{Group and equivariance}
A group is a mathematical structure consisting of a set $G$, which captures the structure of symmetry transformations. The action of a group element $g \in G$ on any function $f \in \gU$ is represented by a transformation $\rmT(.; g )$ as
\bea 
\label{eqn:group_action}
\rmT(f;g) = f(g^{-1} \cdot x),
\eea 
where $g^{-1} \cdot x$ denotes the transformation of $ x $ by the inverse element $ g^{-1} $. In other words,
the function is evaluated at the transformed input $ g^{-1} \cdot x $. Here, $g$ acts as the parameter of the transformation $\rmT$. 
By definition, the transformation needs to satisfy the invertibility condition,
$
\rmT^{-1}(f;g) =  \rmT(f;g^{-1})$, 
for any group element $g$.

Next, a neural network $\gM$ is said to be {\it equivariant} with respect to the group $G$ if $\gM$ commutes with the action of $G$, \ie,  
satisfying the following equality condition
\bea
\label{eqn:def_equ}
\gM( \rmT(f;g)) =  \rmT(\gM(f);g).
\eea
That is, transforming the input similarly transforms the output.
The neural network $\gM$ is {\it invariant} if 
\bea
\label{eqn:def_inv}
\gM( \rmT(f;g)) =  \gM(f),
\eea 
which means transforming the input by a group element $g \in G$ does not change the output.

\myparagraph{Learned canonicalization function} 
A canonicalization function can modify $\gM$, ensuring the resulting $\tilde \gM$ is equivariant. The canonicalization function $h: \gU \to G$ is designed to be equivariant with respect to the group $G$, \ie,
\bea
h(\rmT(f;g)) = g \cdot h(f), \forall g \in G,
\eea  
where $g \cdot h(f)$ denotes a group product.
Using this canonicalization function, the adapted model $\tilde \gM$ is defined as:
\bea 
\label{eqn:canonicalization}
\tilde \gM(f) \triangleq \rmT(\gM {\color{blue}(}\rmT^{-1}(f; \tilde g) {\color{blue})}; \tilde g),
\eea 
where $\tilde g \triangleq h(f)$. In particular, the canonicalization module maps all possible transformed versions of a function $f$ to a canonical input represented as $\rmT^{-1}(f; \tilde g)$. %

The canonicalization function $h$ can be modeled directly by group equivariant networks~\cite{kaba2023equivariance,mondal2023equivariant}. In the absence of an equivariant model, an alternative approach is to formulate the canonicalizer as an optimization problem~\cite{panigrahi2024improved, ma2024canonicalization}
\be
\label{eqn:cann_opt}
h(f) \triangleq \text{arg min}_{g \in G} E( \rmT^{-1}(f;g)).
\ee
Here, $E: \gU \rightarrow \sR$ 
is a learnable \emph{energy function}, \eg, modeled with a neural network.

\section{Latent Deep Equilibrium Canonicalizer}
In~\secref{subsec:loc_scal}, we formulate the monotone scaling group to approximate the local scaling transformation. 
Next, we introduce our DEC module in~\secref{sec:deq_canon} for effective canonicalization on a given module.
Lastly, we propose latent canonicalization for adapting pretrained models in~\secref{sec:laten_canon}. 

\begin{figure}[t]
  \centering
  \small
  \setlength{\tabcolsep}{4pt}
  \renewcommand{\arraystretch}{0.1}
  \begin{tabular}{cc}
    \includegraphics[width=0.45\linewidth, height=0.4\linewidth]{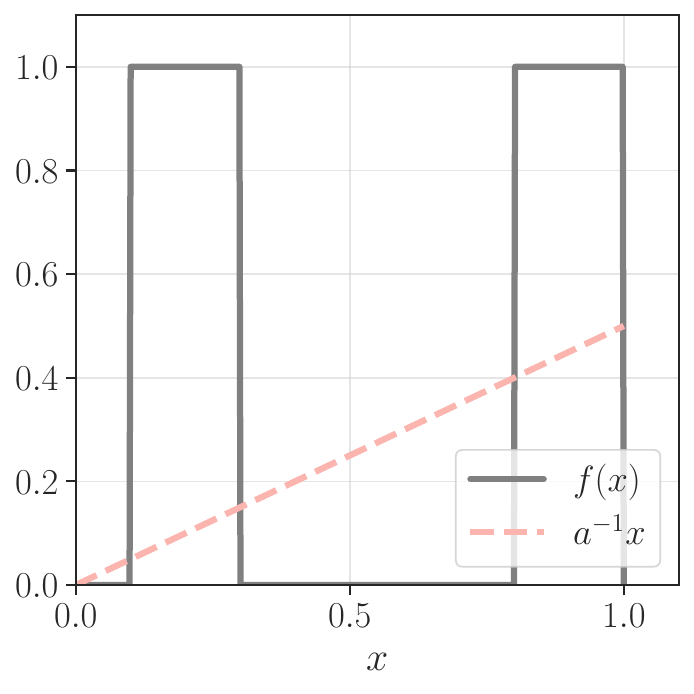} & 
    \includegraphics[width=0.45\linewidth,height=0.4\linewidth]{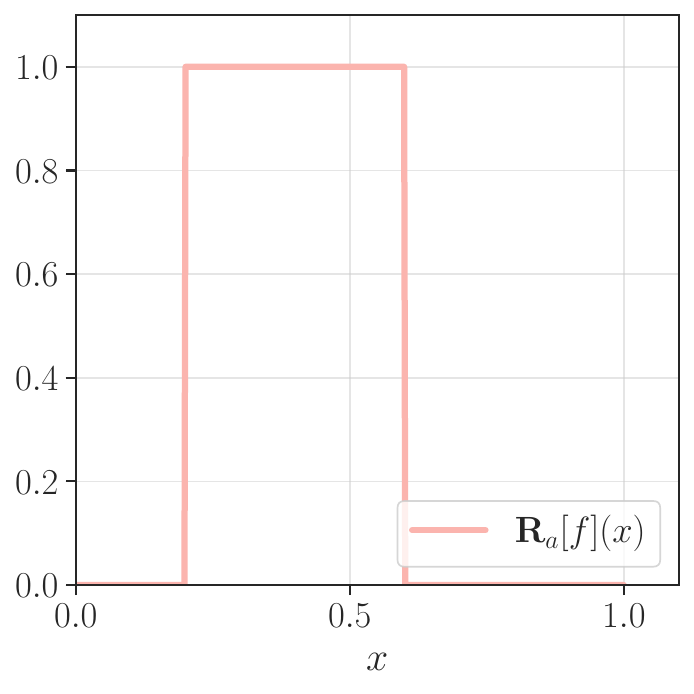} \\
    (a) Global scaling with $a=2.0$. & (b) Globally scaled $\rmR_a[f](x)$. \\
    \\
    \includegraphics[width=0.45\linewidth,height=0.4\linewidth]{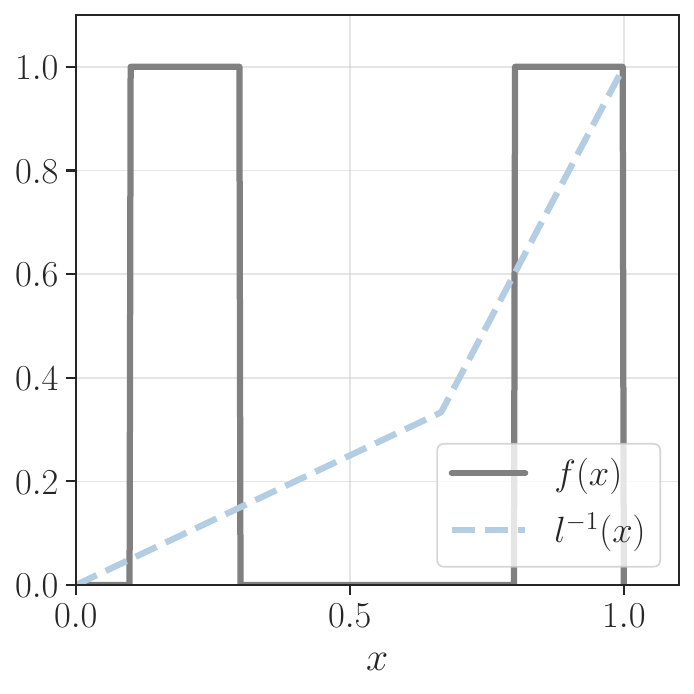} & 
    \includegraphics[width=0.45\linewidth,height=0.4\linewidth]{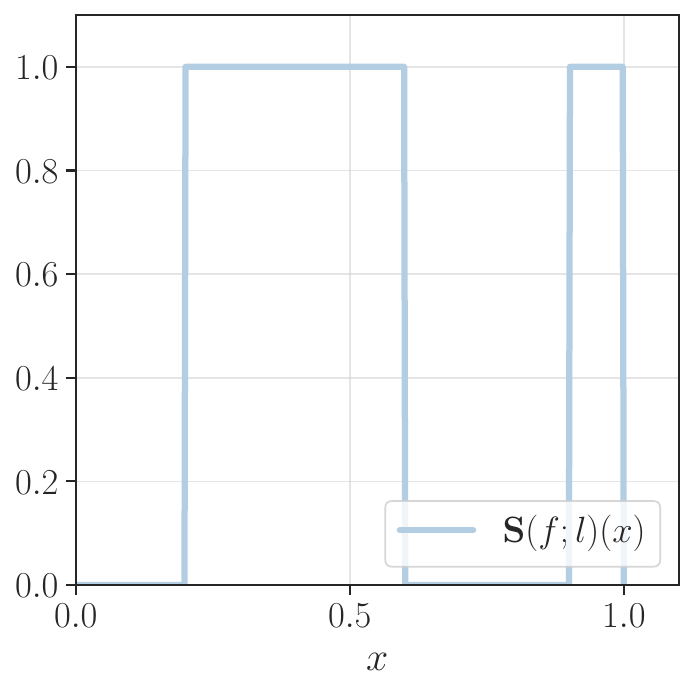} \\
    (c) Local scaling with $l^{-1}(x)$. & (d) Locally scaled $\rmS(f; l)(x)$.
  \end{tabular}
  \vspace{-0.1cm}
  \caption{\textbf{Comparison of global and local scaling.} (a-b) In global scaling, all locations of $f(x)$ are scaled by a constant factor. (c-d) In local scaling, each segment of $f(x)$ is scaled differently.}
  \label{fig:scaling_functions}
  \vspace{-0.55cm}
\end{figure}

\subsection{Local Scaling}
\label{subsec:loc_scal}
In contrast to the global scaling (\equref{eqn:glob_scaling}), scalings in a real-world image are often local. As shown in~\figref{fig:teaser}, different regions in an image have different scaling factors. To approximate such transformations, we propose monotone scaling operations. For readability, the idea is first discussed in 1D and then extended to 2D for images.

\myparagraph{Monotone scaling operation} %
To apply different scaling factors at different parts of the domain, we propose to perform the scaling by a function $l: [0,1] \rightarrow [0,1]$. A monotone scaling is defined as follows 
\bea
\rmS(f;l)(x) \triangleq f(l^{-1}(x)),
\label{eqn:local_scaling_1d}
\eea 
where $l^{-1}$ is the inverse function of $l$ and the local scaling factor at $t$ is $\frac{d l}{dt}$. To preserve the smoothness and invertibility of this transformation, we restrict the functions $l$ to be strictly monotonic, increasing, and continuous. %
In~\figref{fig:scaling_functions}, we provide a comparison between the global and our local scaling operation.
\begin{figure*}[t]
    \centering
    \includegraphics[width=.94\textwidth]{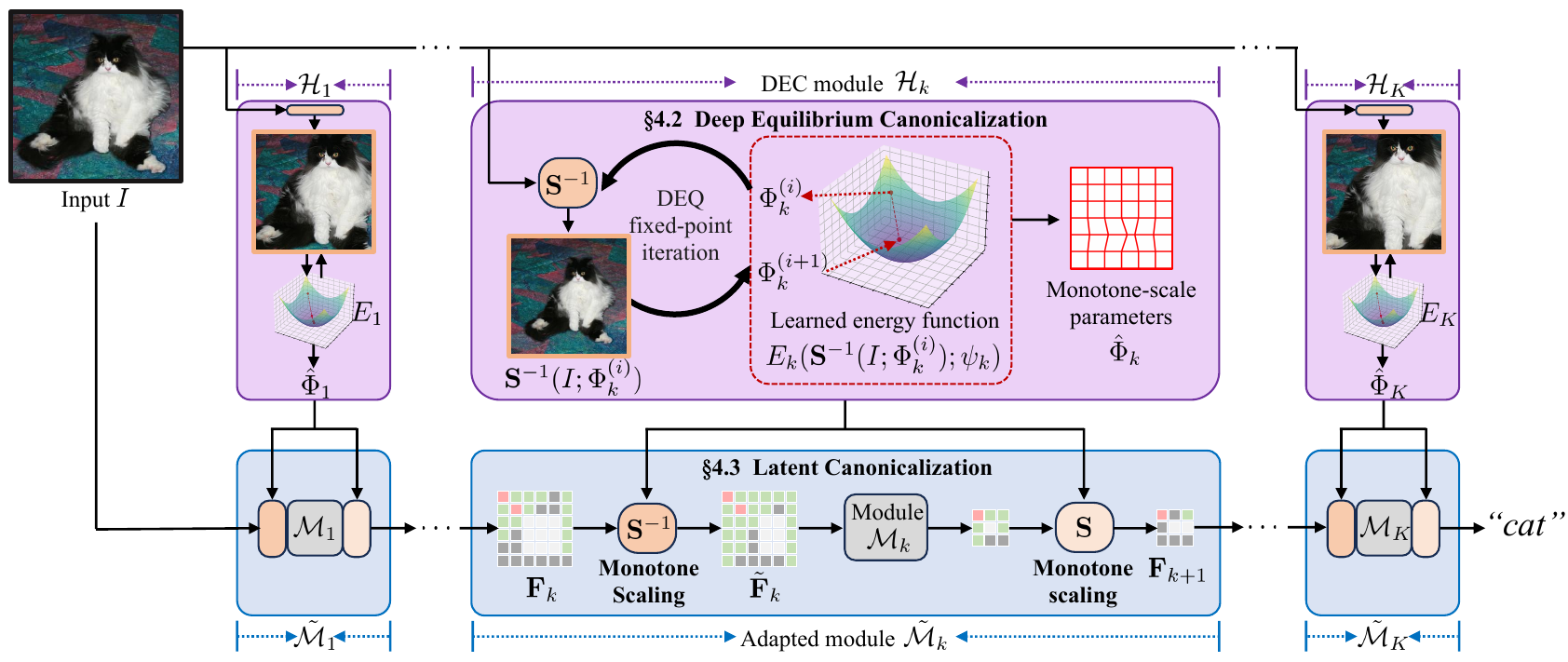} 
        \vspace{-0.2cm}
    \caption{
        \textbf{Illustration of the proposed method.} 
        We formulate a neural network as a sequence of modules $\{ \tilde \gM_k \}_{k=1:K}$.
        We perform latent canonicalization (see \secref{sec:laten_canon}) on the latent feature $\rmF_k$ of each intermediate module. The monotone scaling in the latent canonicalization, \ie, $\rmS^{-1}$ and $\rmS$, are controlled by the parameters $\Phi_k$ found by our DEC module $\gH_k$ (see \ref{sec:deq_canon}).
        Each $\gH_k$ finds the optimal $\hat \Phi_k$ through an iterative fixed-point solver on the energy function $E_k$ until convergence.
    }
    \vspace{-0.3cm}
    \label{fig:pipeline}
\end{figure*}

\myparagraph{Monotone scaling group} 
Denote the set of all continuous and strictly monotonic increasing functions $l$ as $L$. The set $L$ forms a group structure, where the group product $``."$ is represented as the function composition as
\bea
\label{eqn:group_composition}
\left(l_1 \cdot l_2 \right) (x) =(l_1 \circ l_2)(x) =  l_1(l_2(x)).
\eea 
We formally state this in Lemma~\ref{lemma:group_1d}.
\begin{mylemmae}
\begin{restatable}[]{mylemma}{lemmaoned}\label{lemma:group_1d}
The set of all continuous strictly monotonic increasing functions $L$ is a group under the binary operation of function composition.
\end{restatable}
\end{mylemmae}
\begin{proof}
We provide the proof of this lemma in \secref{app:group_proof_id}.
\end{proof}

Lemma~\ref{lemma:group_1d} states that the proposed monotone scaling operation forms a group. Hence, we can formally define equivariance of \wrt the monotone scaling group as
\bea
\label{eqn:loc_sac_eq}
{\tt Equivariance:~} \gM(\rmS(I;l)) = \rmS (\gM(I);l)
\eea
and invariance as
\bea
\label{eqn:loc_sac_in}
{\tt Invariance:~} \gM(\rmS(I;l)) =  \gM(I). 
\eea 
\myparagraph{Parameterization of $l$} The family of possible functions in $l$ is very flexible. For simplicity, we consider a parametric $l$ in the form of a piecewise linear function where the domain $[0,1]$ is discretized uniformly into $N$ discrete points as $\{x_0=0, x_1, \dots, x_N=1\}$. In this case, $l$ has the form 
\bea
l_\Phi(x) =  \phi_{n-1} + \frac{\phi_{n} - \phi_{n-1}}{x_n - x_{n-1}}  \times (x - x_{n-1}),
\eea
when $\Phi = \{\phi_0, \dots, \phi_N\}$ are the parameters. 
We impose the restriction $\phi_{n-1} < \phi_n$ to make the $l$ monotonic increase and further we impose $\phi_0 = 0$ and $\phi_N = 1$ to make the function bijective.
For the ease of notation, we denote the monotone scaling by $l_\Phi$ as $\rmS(f;\Phi)$, by dropping the $l$, \ie, $\rmS(f;\Phi) \triangleq  \rmS(f;l_\Phi)$.

\begin{figure}[t]
    \centering
    \setlength{\tabcolsep}{3pt}
    \begin{tabular}{cc}
    \includegraphics[trim=0 1.8cm 0 0.7cm, clip,width=.45\linewidth, height=.45\linewidth]{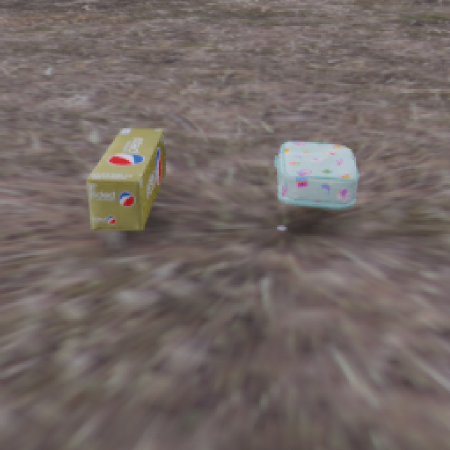} & 
     \includegraphics[trim=0 1.8cm 0 0.7cm, clip,width=.45\linewidth, height=.45\linewidth]{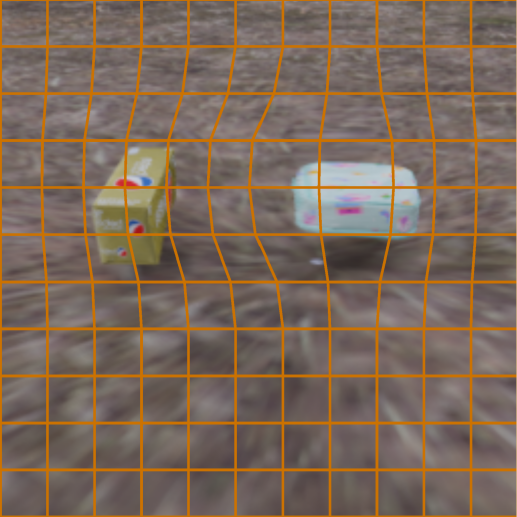}\\
     (a) & (b)\\
    \end{tabular}
    \vspace{-0.2cm}
    \caption{Scaling by monotone function $l:[0,1]^2 \rightarrow [0,1]^2$. (a) The original image. (b) Scaled image. The warped grid visualizes the effect of the monotone scaling operation.
    }
    \vspace{-0.45cm}
    \label{fig:image_loc_scale_2d}
\end{figure}

A similar piecewise linear $l$ can be constructed for monotone scaling operation to images $I: [0, 1]^2 \rightarrow \sR$ by extending the definition of $l$ to the 2D domain, \ie, 
\bea
\label{eqn:local_scaling}
\rmS(I; l)(x,y) = I(l^{-1}(x,y)),
\eea 
where $l: [0,1]^2 \rightarrow [0,1]^2$.
We illustrate this monotone scaling operation in~\figref{fig:image_loc_scale_2d}. %
More details are provided in Appendix~\ref{app:2d_scal_group}. With the parameterization and monotone scaling group defined, we now discuss how to design an effective canonicalizer for this group.

\subsection{Deep Equilibrium Canonicalization (DEC)}
\label{sec:deq_canon}

As reviewed in~\secref{sec:prelim}, one approach to formulate a canonicalizer is by solving for a stationary point, \eg, a minimum of an energy function with respect to the group elements in~\equref{eqn:cann_opt}. However, solving~\equref{eqn:cann_opt} per example and training it end-to-end is slow and memory intensive. Instead, we propose to leverage the idea of amortized optimization~\cite{amos2023tutorial}, \ie, we learn a model to directly predict the solution. In our case, we choose this prediction model to be a Deep equilibrium model to be used as the canonicalizer.

{\vspace*{3pt}\bf\noindent Deep equilibrium model} (DEQ) is a type of implicit neural network where the output is expressed as the fixed/stationary point of a learnable non-linear transformation $\gH$, \ie,
\bea
\label{eqn:deq}
\hat \rvz = \gH(\hat \rvz; I, \psi),
\eea 
where $\gH$ is the non-linear transformation with parameter $\psi$ and $\hat \rvz$ is the  equilibrium hidden state.
The equilibrium can be calculated implicitly by solving the equilibrium condition 
\bea
\rvz - \gH(\rvz;I, \psi) = \vzero
\eea

DEQ can be trained from end to end like any other differentiable module, where the backpropagation can be efficiently computed using implicit differentiation.

\myparagraph{DEC module}
We propose to formulate the canonicalization module using a DEQ. 
That is, we model the optimization in~\equref{eqn:cann_opt}
as a fixed point of a non-linear system $\gH$.  Specifically, we treat $\Phi$ as the hidden state $\rvz$ described in \equref{eqn:deq} and define the update as
\bea
\label{eqn:deq_cano}
 \gH(\Phi^{(i+1)}; I, \psi) \triangleq \Phi^{(i)} - \nabla_{\Phi^{(i)}} E(\rmS^{-1}( I; \Phi^{(i)} ); \psi).
\eea 
As shown in Fig.~\ref{fig:pipeline}, our DEC finds the optimal monotone-scale parameters $\hat \Phi$ via DEQ fixed-point iteration to canonicalize the input $I$.

\begin{myclaime}
\begin{restatable}[]{myclaim}{claimone}
\label{claim:stablepoint}
The optimal monotone-scale parameter, $\hat \Phi$, of energy function $E$ obtained via gradient descent is a fixed point of $\gH$.
\end{restatable}
\end{myclaime}
\begin{proof}
We assume that both DEQ iteration and the gradient descent optimization (\equref{eqn:cann_opt}) start from the same initialization \( \Phi^{(0)} \).
Let $\Phi^{\dag}$ be a fixed point of $\gH$ for the input $I$. Therefore it satisfies
\bea
&& \Phi^{\dag} = \gH( \Phi^{\dag}; I, \psi) \\
& \Rightarrow & \Phi^{\dag} = \hat \Phi^{\dag} - \nabla_{\hat \Phi} E(\rmS^{-1}( I; \Phi^{\dag} ); \psi) \\
&\Rightarrow &\nabla_{\Phi} E(\rmS^{-1}( I; \Phi ); \psi) = \vzero
\eea 
This implies that the gradient of the energy function $E$, $\nabla_{\Phi} E$, evaluated at point $\Phi^{\dag}$ is $\vzero$. Therefore, $\Phi^{\dag}$ corresponds to some stationary point.
Consequently, $\Phi^{\dag}$ corresponds to a local minimum and thus represents the optimal parameter returned by the gradient descent optimization in~\equref{eqn:cann_opt}.
\end{proof}
\vspace{-0.15cm}
The main insight of our canonicalizer design is that, instead of modeling $\gH$ in terms of the gradient of $E$, we directly model $\gH$ using a neural network with learnable parameters $\psi$ and use it as the canonicalization module. Given an image $I$ and monotone-scale parameter \( \Phi^{(i)} \) at step $i$, $\gH$ predict the next-step parameter $\Phi^{(i+1)}$ based on inversely scaled image  $\rmS^{-1}(I, \Phi^{(i)})$. This process is repeated iteratively until convergence to the equilibrium point. The equilibrium point $\hat \Phi$ can be determined by using Anderson acceleration~\cite{anderson1965iterative}.

With our modeled canonicalizer defined, the remaining question is how to incorporate it into an existing deep-net.

\subsection{Latent Canonicalization}
\label{sec:laten_canon}
Typically, canonicalization~\cite{ma2024canonicalization, panigrahi2024improved, mondal2023equivariant} methods apply the canonicalizer at the input of the deep-net. This approach is effective when the transformation of interest is fully represented by the group actions, such as 2D rotation. However, recall that monotone scaling is only an approximation of the real-world local scaling. Therefore, we propose to canonicalize the latent features of an existing network.  We demonstrate the process of latent canonicalization in~\figref{fig:pipeline}. This is motivated by the hypothesis that the inductive bias of monotone scale equivariant features could better approximate an overall local scale equivariant deep-net.

\myparagraph{Canonicalizer on the latent features}
 A deep-net $\gM$ can be viewed as a composition of multiple layers,~\ie,
\bea
\gM(I) = \gM_K \circ \gM_{K-1} \circ \dots \circ \gM_1(I).
\eea
Using this notation, we denote the latent feature maps $\rmF_{k+1} \triangleq \gM_k(\rmF_{k})$ where the input feature is the image, \ie, $\rmF_1 = I$.
For each layer $\gM_k$, we use canonicalization to adapt it to $\tilde \gM_k$ %
following~\equref{eqn:canonicalization}. 
Our overall adapted model with equivariance guarantee is a composition of all adapted modules, \ie $\tilde \gM = \tilde \gM_K \dots \circ \tilde \gM_0$.

For monotone scale equivariance, 
we define each adapted module $\tilde\gM_k$ as follows:
\bea
\label{eqn:laten_canon}
\tilde \gM^{\tt Eq}_k(\rmF_k) = \rmS(\gM_k(\tilde \rmF_k);\Phi_k),
\eea where $\tilde \rmF_k = \rmS^{-1}(\rmF_k; \Phi_k))$.
That is, we first apply monotone scaling $\mS^{-1}$ to each input feature of each layer, \ie, $\rmF_k$, and subsequently apply its inverse $\mS$ to the output of $\tilde \gM_k$. This ensures that the latent feature is equivariant to monotone scaling.
Similarly, for monotone scale invariance, we define the adapted modules as
\bea 
\tilde \gM^{\tt Inv}_k(\rmF_k) = \gM_k(\tilde \rmF_k).
\eea

\section{Experiments}
For evaluation, we aim to benchmark the local scale consistency of existing architectures and verify that the proposed DEC can improve consistency while maintaining model performance. To accurately measure the local scale consistency, we create datasets of locally scaled images where we know the underlying local scale transformation applied to each image. Specifically, we considered semantic segmentation on a dataset based on the Google Scan Objects~\cite{downs2022google} and image classification using the MNIST~\cite{lecun1998mnist} dataset and the ImageNet~\cite{deng2009imagenet} dataset.  
The implementation details are in the Appendix~\ref{app:supp_impl}.
More results are provided in the Appendix~\ref{app:results}.

\subsection{Semantic Segmentation}\label{sec:seg}
\begin{table}[t]
    \centering
    \setlength{\tabcolsep}{.2mm}
    \renewcommand{\arraystretch}{0}
    \begin{tabular}{cc@{\hskip 8pt}cc}
        \includegraphics[width=0.23\linewidth]{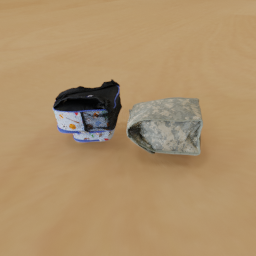} &
        \includegraphics[width=0.23\linewidth]{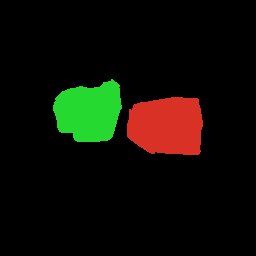} &
        \includegraphics[width=0.23\linewidth]{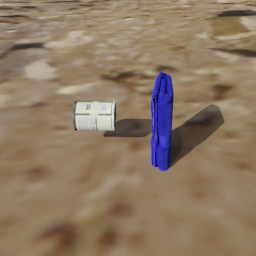} &
        \includegraphics[width=0.23\linewidth]{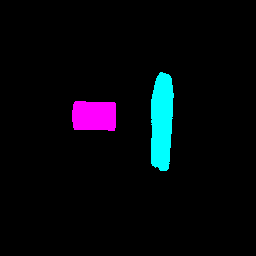}
        \\
        \vspace{1pt} \\
        \includegraphics[width=0.23\linewidth]{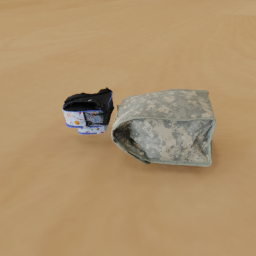} &
        \includegraphics[width=0.23\linewidth]{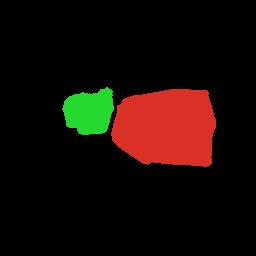} &
        \includegraphics[width=0.23\linewidth]{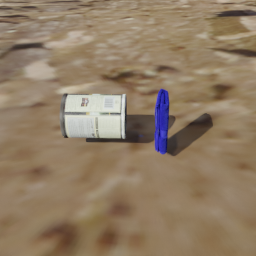} &
        \includegraphics[width=0.23\linewidth]{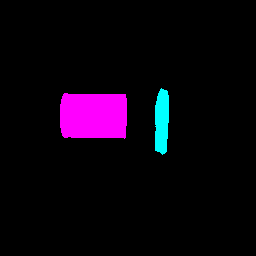} \\
    \end{tabular}
    \captionof{figure}{
         \textbf{Samples from locally scaled object segmentation dataset.} Each object in an image is scaled independently.
    }
    \vspace{-0.15cm}
    \label{fig:syn_segmen}
\end{table}

\begin{table}[t]
\centering
\small
\setlength{\tabcolsep}{3pt}
    \begin{tabular}{lcccccc}
    \toprule
    \multirow{2}{*}{\textbf{Method}} & \multicolumn{2}{c}{{\bf ViT}~\cite{dosovitskiy2020image}} & \multicolumn{2}{c}{{\bf Swin}~\cite{liu2021swin}} & \multicolumn{2}{c}{{\bf DINOv2}~\cite{oquab2023dinov2}}
    \\ 
    & {mIoU}$\uparrow$ & {EquE}$\downarrow$ 
    & {mIoU}$\uparrow$ & {EquE}$\downarrow$ 
    & {mIoU}$\uparrow$ & {EquE}$\downarrow$ 
    \\
    \midrule
    \baselineA & 82.18 & 0.054 & 86.65      & 0.036 & 76.63  & 0.061
    \\
    \midrule
    \baselineB & 79.37   & 0.062 & 83.94       & 0.050 & 76.94  & 0.064
    \\
    \baselineD & 82.35   & 0.054 & 87.63       & 0.037 & 79.05  & 0.058
    \\
    \ourrow Ours     & \bf 83.12   & \bf 0.052 & \bf 88.56       & \bf 0.035 & \bf 80.86  & \bf 0.055\\
    \bottomrule
    \end{tabular}
\caption{
    \textbf{Results on local scale equivariance.} For each method, we report the mIoU(\%) and EquE across 3 architectures.
}
\vspace{-0.5cm}
\label{tab:lsos_results}
\end{table}

\myparagraph{Dataset setup} For evaluating equivariance, we construct a locally scaled segmentation dataset by simulating real-world objects' scale variations. To have precise control over the variations, we render objects from Google Scanned Objects~\cite{downs2022google} on a randomly selected background in HDRI Haven~\cite{zaal2020hdri}. For each image, we place two 3D objects and extract the ground-truth mask. 

To introduce the local scaling effects, we vary the relative distance of each object from the camera and enable independent scale changes for each object while keeping the background fixed. The dataset includes 100 distinct objects; see~\figref{fig:syn_segmen} for examples. We use 6,000 images for training and 2,000 for testing with an image resolution of $256 \times 256$.

\myparagraph{Evaluation metric} To measure performance, we report the mean intersection over union (mIoU). To measure consistency, we propose monotone scale equivariance error (\text{EquE}) on the predicted masked, defined as
\bea\label{eq:eque}
\frac{1}{|\gD|} \sum_{I \in \gD} \E_{_{\Phi \sim \Omega}} \|\rmS(\gM(I);\Phi) - \gM(\rmS(I; \Phi))\|_2^2
\eea 
over a dataset $\gD$ and $\Omega$ denotes a uniform distribution over all possible monotone scale parameters.

Intuitively, EquE can be thought of as a {\it soft} definition, in the $\ell_2$ sense, of the monotone scale equivariance in~\equref{eqn:loc_sac_eq}. That is, an equivariant model will achieve an EquE of zero.

\myparagraph{Baselines} We consider the following alternative methods to encourage local scale consistency into a model:
\begin{itemize}
    \item \baselineA: We train the architecture with random monotone scaling data augmentation on the training set. \baselineA~also serves as the starting weights for DEC and the other baselines.
    
    \item \baselineB: We discretize the local scaling parameter and consider a set of $64$ different parameters, including the identity transformation. We use a $3$-layer CNN to model the energy function in~\equref{eqn:cann_opt}.
    
    \item \baselineD: As one of the metrics is the equivariance error, we further fine-tune \baselineA~by using EquE in~\equref{eq:eque} as part of the loss to encourage local scale equivariance.
\end{itemize}
The baselines and our DEC are applied to three deep-net architectures: ViT~\cite{dosovitskiy2020image}, Swin~\cite{liu2021swin},  and DINOv2 \cite{oquab2023dinov2}. 

\begin{table}[t]
    \centering
    \setlength{\tabcolsep}{1pt}
    \renewcommand{\arraystretch}{0}
    \begin{tabular}{ccc}
        \includegraphics[width=0.33\linewidth, height=0.33\linewidth]{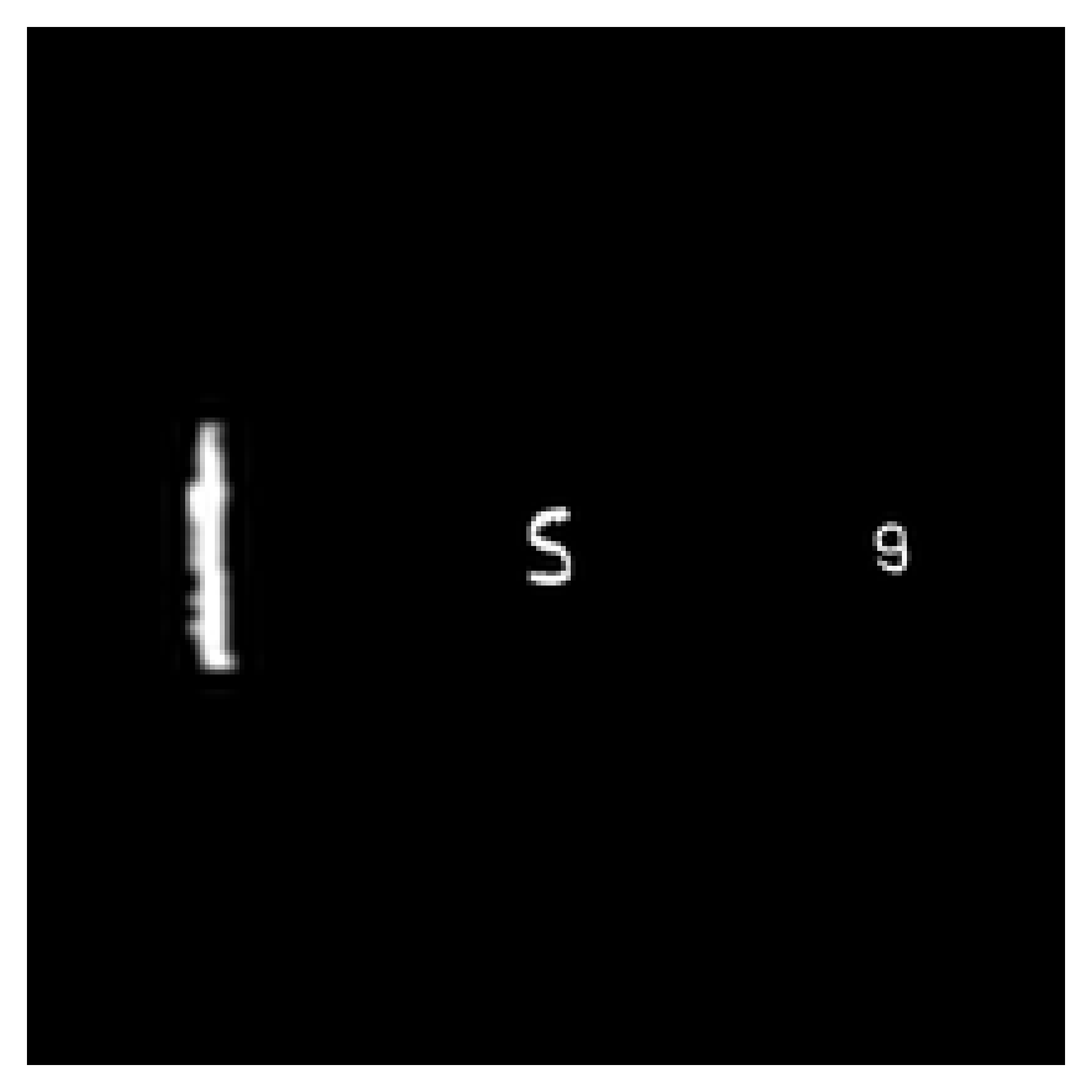} &
        \includegraphics[width=0.33\linewidth, height=0.33\linewidth]{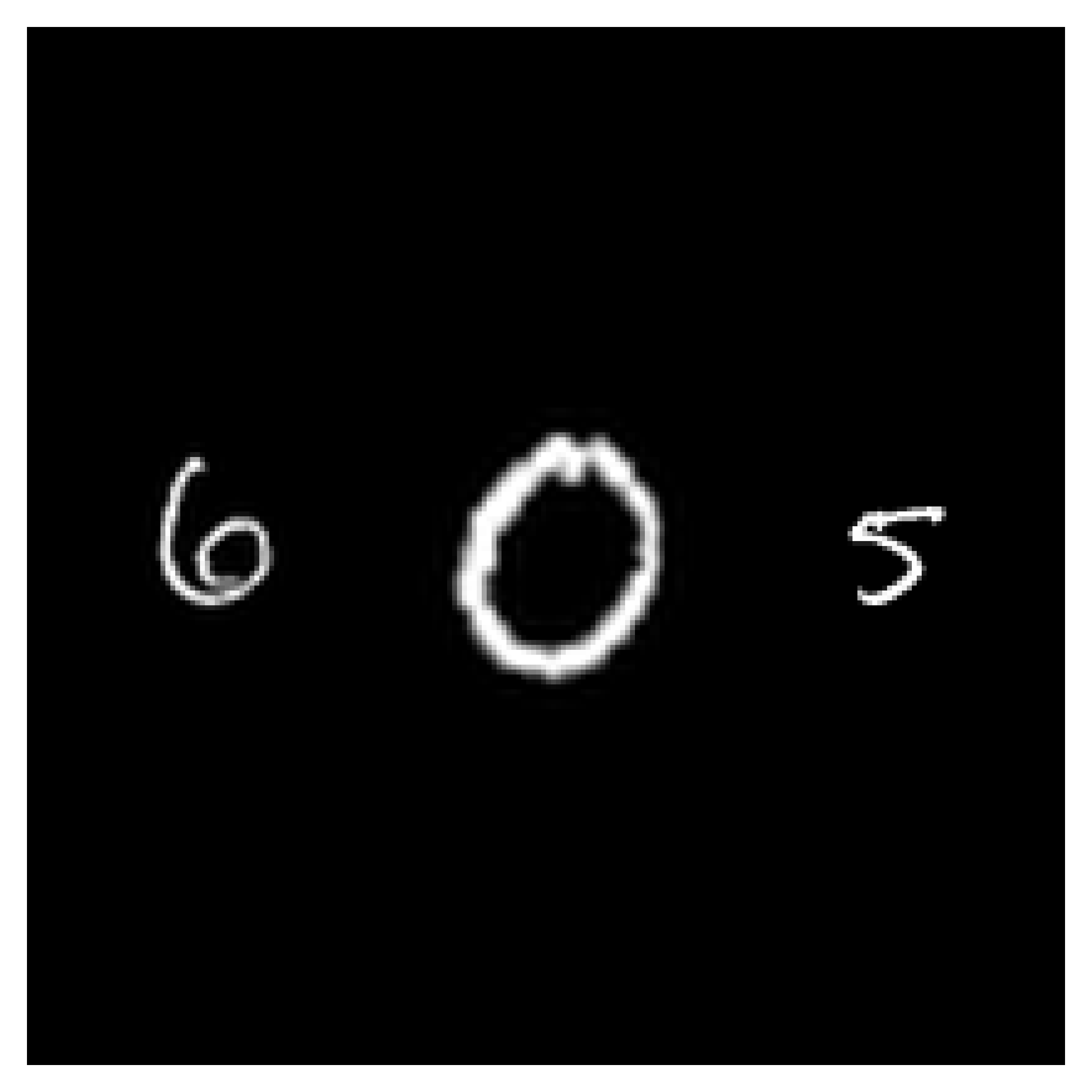} & 
        \includegraphics[width=0.33\linewidth, height=0.33\linewidth]{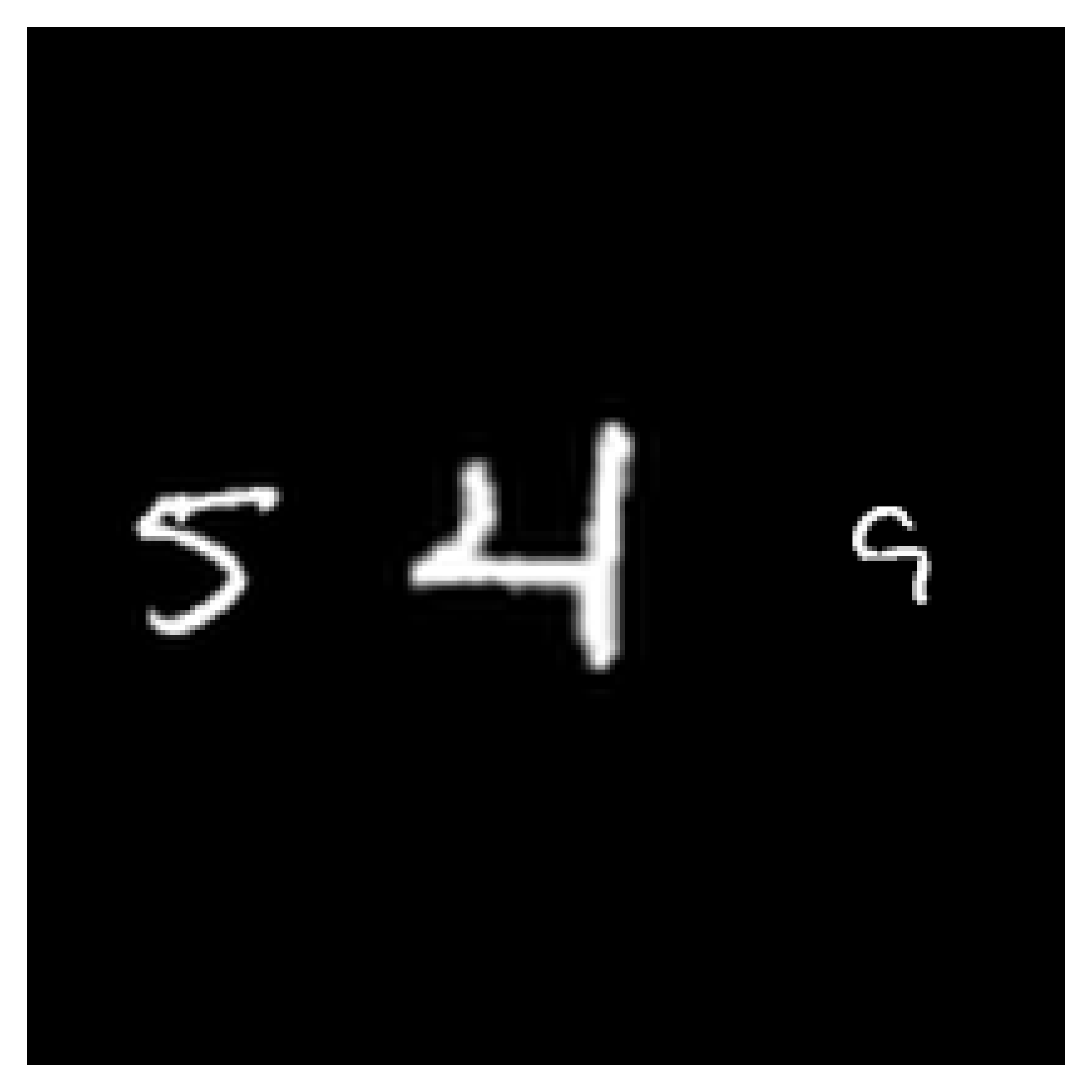}
        \\
    \end{tabular}
    \captionof{figure}{
         \textbf{Samples from locally scaled MNIST.} The task is to regress the number,~\ie, ``159'', ``605'', and ``549'' are the ground-truth for each of the images. In other words, an image classification task of 1000 classes; from 0 to 999.}
    \label{fig:mnist}
    \vspace{-0.35cm}
\end{table}

\begin{table*}[t]
\centering
\small
\setlength{\tabcolsep}{2.5pt}
\resizebox{1.01\textwidth}{!}{
    \begin{tabular}{l
    cr 
    cr 
    cr 
    cr 
    cr 
    cr
    }
    \toprule
    \multirow{2}{*}{\textbf{Method}} & 
    \multicolumn{2}{c}{\textbf{ResNet}~\cite{he2016resnet}} & 
    \multicolumn{2}{c}{\textbf{ViT}~\cite{dosovitskiy2020image}} &
    \multicolumn{2}{c}{\textbf{DeiT}~\cite{touvron2021deit}} &
    \multicolumn{2}{c}{{\bf Swin}~\cite{liu2021swin}} & 
    \multicolumn{2}{c}{\textbf{BEiT}~\cite{bao2021beit}} & 
    \multicolumn{2}{c}{\textbf{DINOv2}~\cite{oquab2023dinov2}}
    \\
    & Acc$\uparrow$ & {InvE$\downarrow$}
    & Acc$\uparrow$ & {InvE$\downarrow$}
    & Acc$\uparrow$ & {InvE$\downarrow$}
    & Acc$\uparrow$ & {InvE$\downarrow$}
    & Acc$\uparrow$ & {InvE$\downarrow$}
    & Acc$\uparrow$ & {InvE$\downarrow$}
    \\
    \midrule
    \baselineA & 93.09 $\pm$0.41 & 13.71 $\pm$ 1.65 & 91.23 $\pm$ 1.80 & 13.33 $\pm$ 1.72 &90.91 $\pm$ 1.18  & 16.17 $\pm$ 4.94 & 94.26 $\pm$ 0.25  & 5.44 $\pm$ 1.65 & 94.70 $\pm$ 0.37 & 9.69 $\pm$ 1.94 & 95.80 $\pm$ 0.29 & 7.47 $\pm$ 1.33
    \\ 
    \midrule
    \baselineB & 94.22 $\pm$ 0.33 & 13.84 $\pm$ 1.05 & 93.73 $\pm$ 0.61 & 12.32 $\pm$ 2.35 &
    95.01 $\pm$ 0.20 & 7.04 $\pm$ 0.70 &
    95.67 $\pm$ 0.19 & 4.27 $\pm$ 1.86 &
    95.92 $\pm$ 0.29 & 5.07 $\pm$ 1.65 &
    96.63 $\pm$ 0.43  & 6.00 $\pm$ 3.12
    \\
    \baselineC & 93.67 $\pm$ 0.73 & 13.47 $\pm$ 2.01 & 94.79 $\pm$ 0.42 & 8.17 $\pm$ 2.60 &
    94.06 $\pm$ 0.38 & 8.04 $\pm$ 2.88 &
    94.78 $\pm$ 0.25 & 3.89 $\pm$ 0.62 & 
    95.26 $\pm$ 0.41 & 6.49 $\pm$ 0.78 &
    96.66 $\pm$ 0.10  & 5.21 $\pm$ 1.34
    \\    
    \ourrow Ours & \bf 96.09 $\pm$ 0.16 & {\bf 6.13 $\pm$ 2.01} & \bf 95.64 $\pm$ 0.15 & {\bf 5.04 $\pm$ 3.37} &
    \bf 96.46 $\pm$ 0.14 & {\bf 6.02 $\pm$ 2.71} &
    \bf 96.91 $\pm$ 0.12 & {\bf 2.08 $\pm$ 1.09} & 
    \bf 97.05 $\pm$ 0.26 & {\bf 4.13 $\pm$ 1.70} & 
    \bf 97.90 $\pm$ 0.21 & {\bf 2.62 $\pm$ 1.69}    
    \\
    \bottomrule
    \end{tabular}
}
\caption{
    \textbf{Results of local scale invariance on MNIST.} For each method, we report the Acc($\%$) and InvE($10^{-2}$) across 6 architectures, with error bars computed over five runs.
}
\vspace{-0.4cm}
\label{tab:mnist_results}
\end{table*}

\begin{table}[t]
\centering
\small
    \begin{tabular}{lcccc}
    \toprule
    \textbf{Scale} &{\bf $[0.4,1.0]$} & {\bf $[1.0,2.0]$} &{\bf $[2.0,3.0]$} & Overall\\
    \midrule
    \baselineA & 91.64 & 96.32   & 93.84  & 93.93 $\pm$ 1.91\\
    \midrule
    \baselineB & 90.16  & 95.39  & 90.54   & 	92.03 $\pm$ 2.38\\
    \baselineC & 91.98  & 96.26  & 93.88   & 	94.04 $\pm$ 1.75\\
    \ourrow Ours & \bf 95.14 & \bf 97.81  & \bf 96.64  & 	\bf 96.53 $\pm$	1.09\\
    \bottomrule
    \end{tabular}
\caption{
    \textbf{Per-scale results on MNIST.} We report the accuracy across various local scaling ranges using the Swin~\cite{liu2021swin} architecture.
}
\vspace{-0.4cm}
\label{tab:per_scale_mnist}
\end{table}

\myparagraph{Results} 
In \tabref{tab:lsos_results}, we compare our DEC with the equivariant adaptation baselines in semantic segmentation. We observe that ours achieves the highest mIoU and lowest EquE in all cases. Although \baselineA~is effective in pursuing local scale equivariance, they are far from optimal. Comparison of \baselineA~and ours shows that mere data augmentation with local scaling operations is not reliable enough to guarantee equivariance, nor does further guiding \baselineA~with the~\equref{eq:eque} in \baselineD.

In contrast, \baselineB~degrades the performance of \baselineA~in most cases. That is, complete canonicalization on the input image only induces deterioration to the model. This shows the importance of canonicalization on latent features and replacing gradient optimization with our DEC,~\ie, solving a fixed-point equation iteratively.

Lastly, consistent improvements of ours on 3 distinct architectures demonstrate the generalizability of our model to different deep-nets.
On average, ours outperforms the best baseline, \ie, \baselineD, in mIoU by 1.17\%.

\subsection{Image Classification --- MNIST}
\myparagraph{Dataset setup} 
Inspired by the MNIST-scale dataset from scale equivariance~\cite{kanazawa2014locally,sosnovik2021disco,rahman2024truly}, 
we construct a locally scaled MNIST by creating 3-digit images from ``000'' to ``999''. Each digit is randomly resized by a factor within $[0.4, 2.0]$, allowing for a scale variation of up to $5$ times between different digits (see~\figref{fig:mnist}). The images have a resolution of $224 \times 224$. This provides a suitable setup to test the local scale invariance, where the task is to predict the 1,000 number classes. As this task is relatively easy, we consider the more difficult setting with limited training data. The training set consists of 6,000 samples, while the test set contains 50,000 samples. The train/test images of the locally scaled dataset are sampled from the train/test split of the MNIST, respectively.

\myparagraph{Evaluation Metric}
We report the top-1 classification accuracy and the local scale invariance error
\bea
\label{eqn:inv_err}
\text{InvE} \triangleq \frac{1}{|\gD| |\gS_I|} \sum_{I \in \gD} \sum_{I_s \in \gS_I} {||\gM(I) - \gM(I_s)||_2^2}{ }. 
\eea 
Here, $\gS_I$ consists of all variants of locally scaled $I$. 
Intuitively, \text{InvE} can be thought as a {\it soft} definition, in the $\ell_2$ sense, of the local scale invariance in~\equref{eqn:loc_sac_in}.

\myparagraph{Baselines} We follow the same baselines of \baselineA\ and  \baselineB\ from~\secref{sec:seg}. Differently, as image classification should be invariant to local scaling, we instead have the baseline of InvL, which contains an additional loss of
\bea 
    \label{eqn:inv_loss}
    \gL_{\tt Inv} = \E_{I \sim \gD, \Phi \sim \Omega} || \gM(I) - \gM(\rmS(I;\Phi)||_2^2,
\eea 
to encourage monotone scale invariance.
We evaluate these baselines on commonly used image classification architectures, including 
ResNet~\cite{he2016resnet}, ViT~\cite{dosovitskiy2020image}, DeiT~\cite{touvron2021deit}, Swin \cite{liu2021swin}, BEiT~\cite{bao2021beit}, and DINOv2~\cite{oquab2023dinov2}.

\myparagraph{Results} In~\tabref{tab:mnist_results}, we compare our DEC with the baselines in image classification on MNIST. 
We have observations similar to the results in~\tabref{tab:lsos_results}.
\baselineA~and \baselineC~are not optimal while \baselineD~degrades the performance of some architectures, \eg ResNet, ViT, and Swin. Ours achieves the highest Acc and lowest InvE, no matter the architectures.

In particular, ours shows impressive results on local scale invariance and achieves a significant improvement in InvE compared to \baselineA~and other baselines. Specifically, ours reduces InvE from 18.08 to 9.85 on ResNet and from 6.30 to 1.31 on DINOv2. This substantiates DEC's ability to achieve local scaling invariance. On average, ours improves \baselineA~on Acc by 4.10\% and InvE by 8.91$\times 10^{-2}$. To further demonstrate the broad applicability of our model, we provide an additional comparison against hierarchical models \cite{tian2023resformer, gu2022multi} in \secref{sec:additional_baselines}.

\myparagraph{Analysis and ablation studies}
In~\tabref{tab:per_scale_mnist}, we present the baseline results on Swin~\cite{liu2021swin} across different local scale ranges. Specifically, we restrict the local scale range of each number digit in various scale brackets. We note that our method achieved higher accuracy in every scale range. Furthermore, by accessing different scale ranges, our method demonstrates the most consistent accuracy, reducing the std to $1.09$ from $1.91$. Additional ablations on DEC-related hyperparameters are provided in \secref{sec:additional_ablation}.

We also compare the time and memory requirements of DEC modules with differentiable optimization-based canonicalization, \ie,~{\it Optim}. We train both methods with a batch size of 10 and image sizes of $224 \times 224$. {\it Optim} takes 43.30 GB of GPU memory and spends 0.41s per iteration, while our DEC takes 5.75 GB and spends 0.19s. In other words, {\it Optim} requires more than $8$ times the memory and twice the time than the DEC module.
As discussed in~\secref{sec:deq_canon}, this makes it extremely difficult to apply {\it Optim} to deeper networks. We provide further discussion on runtime in \secref{app:runtime}.

\begin{table}[t]
    \centering
    \setlength{\tabcolsep}{1.2pt}
    \renewcommand{\arraystretch}{0.5}
    \begin{tabular}{cccc}
        \includegraphics[width=0.24\linewidth, height=0.24\linewidth]{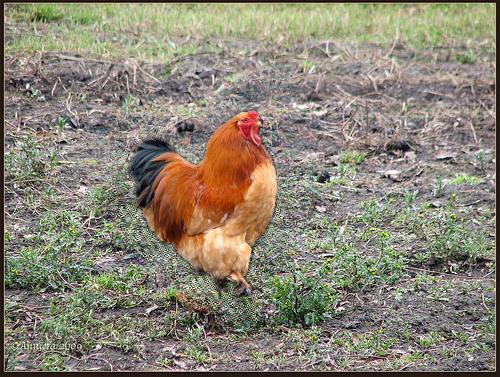} &
        \includegraphics[width=0.24\linewidth, height=0.24\linewidth]{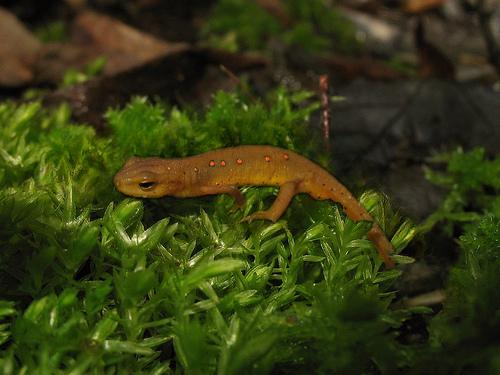} &
        \includegraphics[width=0.24\linewidth, height=0.24\linewidth]{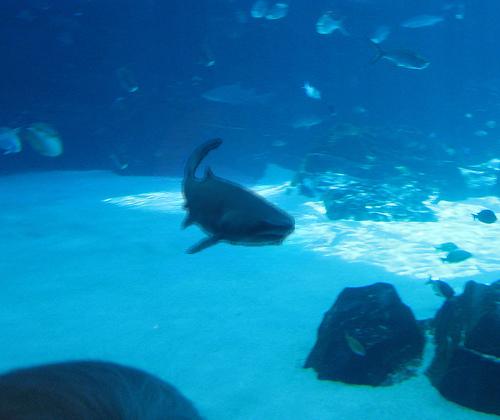} &
        \includegraphics[width=0.24\linewidth, height=0.24\linewidth]{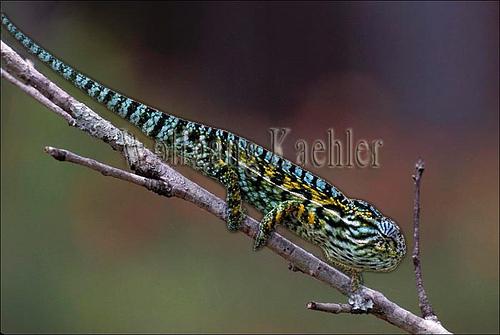}
        \\
        \includegraphics[width=0.24\linewidth, height=0.24\linewidth]{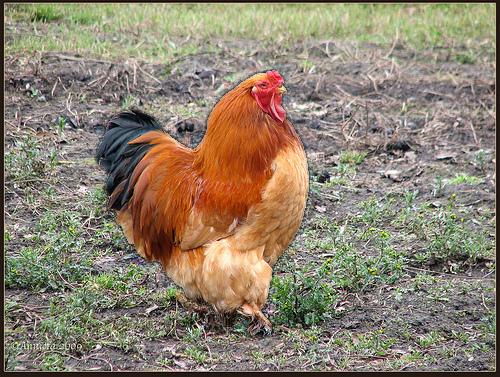} &
        \includegraphics[width=0.24\linewidth, height=0.24\linewidth]{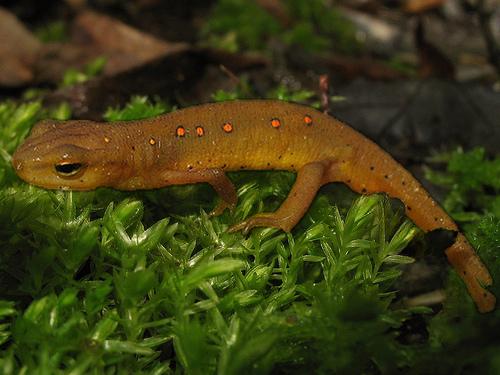} & 
        \includegraphics[width=0.24\linewidth, height=0.24\linewidth]{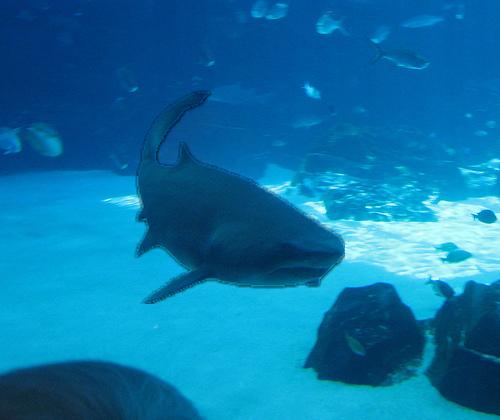} &
        \includegraphics[width=0.24\linewidth, height=0.24\linewidth]{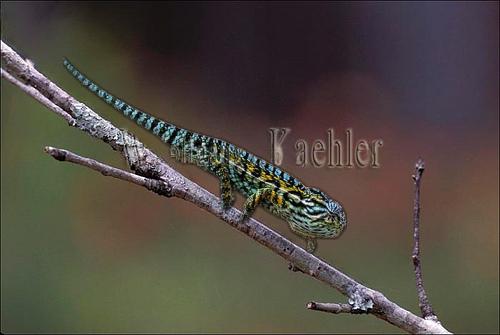} \\
    \end{tabular}
    \captionof{figure}{
         \textbf{Samples from locally scaled ImageNet.} For each image, we show the target object at two different scales. Note that the scale of the background is always unchanged.
    }
    \vspace{-0.5cm}
    \label{fig:scale_imn}
\end{table}

\begin{table*}[t]
    \small
    \centering
    \setlength{\tabcolsep}{3pt}
    \renewcommand{\arraystretch}{1}
    \begin{tabular}{cccccc}
    \includegraphics[width=0.16\linewidth, height=0.16\linewidth]{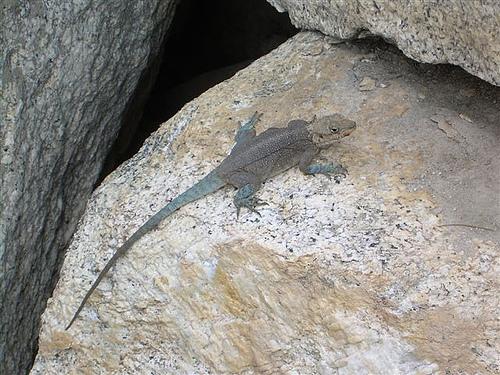} &
    \includegraphics[width=0.16\linewidth, height=0.16\linewidth]{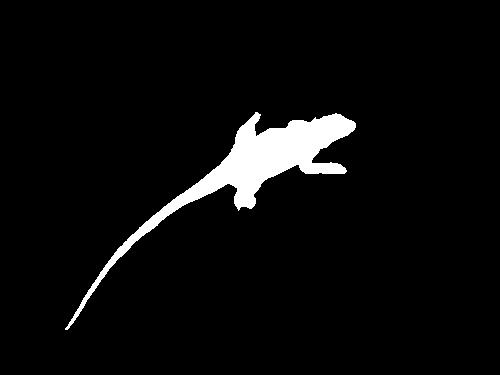} &
    \includegraphics[width=0.16\linewidth, height=0.16\linewidth]{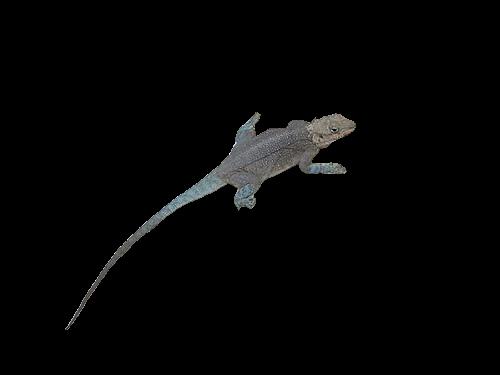} &
    \includegraphics[width=0.16\linewidth, height=0.16\linewidth]{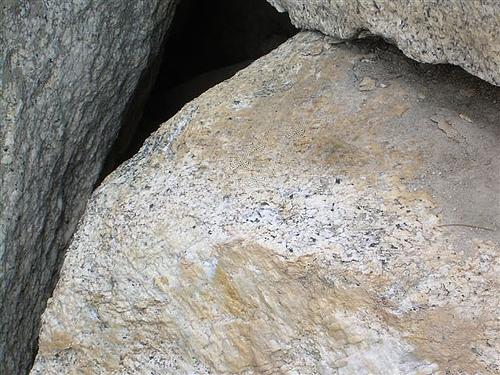} &
    \includegraphics[trim=50 30 55 50, clip, width=0.16\linewidth, height=0.16\linewidth]{figure/inpain_pipeLine/n01687978_7485_extracted_object_.JPEG} &
    \includegraphics[width=0.16\linewidth, height=0.16\linewidth]{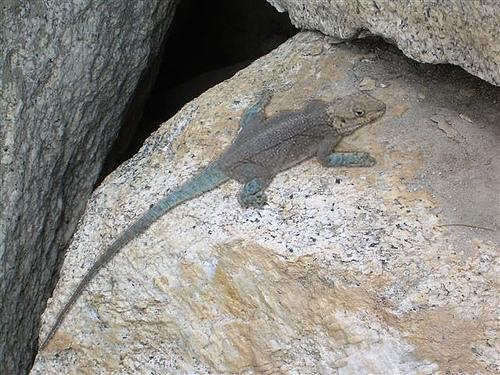} \\
    (a) Original Image & (b) Object Mask & (c) Extracted Object &
    (d) Inpainted Image & (e) Scaling ($\times 1.2$) & (f) Locally Scaled \\
\end{tabular}
    \vspace{-0.18cm}
    \caption{
         \textbf{Generation procedure of locally scaled ImageNet.}  
         Given (a) an image of a lizard, we first (b) obtain a mask of the lizard and (c) extract the lizard from the image. We then (d) inpaint the region without the lizard to get the background. Finally, (e) we scale the lizard by $1.2\times$ and (f) place it back onto the background to get the locally scaled image. %
    }
    \label{fig:inpaint_pipeline}
        \vspace{-0.3cm}
\end{table*}

\subsection{Image Classification --- ImageNet}
\myparagraph{Data setup} This experiment aims to benchmark local scale invariance in a more realistic real-world setting, where we create locally scaled images using ImageNet~\cite{deng2009imagenet}. As shown in~\figref{fig:scale_imn}, only the target object in each image is scaled, while the background is kept the same.

In more detail, these images are created using the following procedure (see~\figref{fig:inpaint_pipeline} for visualization): (a) We extract the bounding box of the target object by Grounding DINO~\cite{liu2024grounding} using the image labels. (b) With the bounding box, we extract a fine object mask using SAM~\cite{kirillov2023segment} and (c) extract the object. (d) We inpaint the object mas region using LAMA~\cite{suvorov2022resolution}. (e) We scale the extracted object and (f) place the scaled object back onto the inpainted background. The object of interest is scaled by a factor within the range of $[0.7, 1.3]$. %
We fine-tune the pretrained models on a training dataset containing 10,000 images. We use 50,000 testing images for evaluation. 

\myparagraph{Baselines} 
For ImageNet, we use open-source models from Pytorch Image Models~\cite{rw2019timm}. We fine-tune the pretrained weights using their training script, which we denote as \generic. All other baselines are initialized from \generic.
We consider four different architectures ViT~\cite{dosovitskiy2020image}, DeiT~\cite{touvron2021deit}, Swin~\cite{liu2021swin}, and BEiT~\cite{bao2021beit}.
During fine-tuning, all baselines follow the best practices of data augmentation and fine-tuning strategy for vision transformers~\cite{he2022masked, liu2021swin}, including techniques such as label-smoothing, cutmix, \etc.

\begin{table}[t]
\centering
\small
\setlength{\tabcolsep}{2.5pt}
\resizebox{\linewidth}{!}{
    \begin{tabular}{l
    cr c
    cr c
    cr c
    cr}
    \toprule
    \multirow{2}{*}{\textbf{Method}} & \multicolumn{2}{c}{\bf ViT} && 
    \multicolumn{2}{c}{\textbf{DeiT}} && \multicolumn{2}{c}{\textbf{Swin}} && \multicolumn{2}{c}{\textbf{BEiT}} \\
    & Acc$\uparrow$ & {InvE$\downarrow$}
    && Acc$\uparrow$ & {InvE$\downarrow$}
    && Acc$\uparrow$ & {InvE$\downarrow$}
    && Acc$\uparrow$ & {InvE$\downarrow$}
    \\
    \midrule
    \generic   & 80.15  & 8.18 && 69.06  & 11.13 &&  77.94 & 8.93 && 84.70  & 6.29\\
    \baselineA & 80.15  & 8.16 && 69.06  & 11.12 && 77.96 & 8.92 && 84.66  & 6.27
    \\        
    \midrule
    \baselineB & 77.72  & 10.20 && 64.79  & 12.74 && 74.67 & 11.40 && 83.34  & 8.03
    \\
    \baselineC & 80.16  & 8.16 && 69.06  & 11.12 && 77.97 & 8.92 && 84.66  & 6.27
    \\
    \ourrow Ours & \bf 80.36 & \bf 8.10
    && \bf 69.27 & \bf 11.08 && \bf 78.32 & \bf 8.82 && \bf 85.08 & \bf 6.24
     \\
    \bottomrule
    \end{tabular}
}
\vspace{-0.1cm}
\caption{
    \textbf{Results of local scale invariance on ImageNet.} For each method, we report the Top-1 Acc($\%$) and InvE($10^{-2}$) across 4 architectures.
}
\label{tab:imagenet_results}
\vspace{-0.1cm}
\end{table}

\begin{table}[t]
\centering
\small
\setlength{\tabcolsep}{2.5pt}
\resizebox{\linewidth}{!}{
    \begin{tabular}{l
    cccccccc}
    \toprule
    \multirow{2}{*}{\textbf{Method}} & \multicolumn{8}{c}{\bf Local Scales}  \\
    & $0.7$
    & $0.8$
    & $0.9$
    & $1.0$
    & $1.1$
    & $1.2$
    & $1.3$
    & Overall
    \\
    \midrule
    \generic   & 74.66 & 76.44 & 78.07 & 79.55 & 79.48 & 78.88 & 78.53 & 77.94 $\pm$ 1.65
    \\
    \baselineA & 74.71 & 76.47 & 78.07 & 79.56 & 79.48 & 78.87 & 78.57 & 77.96 $\pm$ 1.64
    \\        
    \midrule
    \baselineB & 71.37 & 73.28 & 74.87 & 76.04 & 76.04 & 75.79 & 75.32 & 74.67 $\pm$ 1.61
    \\
    \baselineC & 74.73 & 76.46 & 78.07 & 79.58 & 79.48 & 78.87 & 78.57 & 77.97 $\pm$ 1.64
    \\
    \ourrow Ours & 
    \bf 75.31 & \bf 77.04 & \bf 78.65 & \bf 79.86 & \bf 79.60 & \bf 79.05 & \bf 78.73 & {\bf 78.32} $\pm$ {\bf 1.49}
     \\
    \bottomrule
    \end{tabular}
}
\vspace{-0.15cm}
\caption{
    \textbf{Per-scale results on ImageNet.}  We report the Top-1 Acc($\%$) across various local scaling factors.  Swin~\cite{liu2021swin}.
}
\vspace{-0.4cm}
\label{tab:per_scale_imnet}
\end{table}

\myparagraph{Results} 
We present the ImageNet results  in~\tabref{tab:imagenet_results}. We observe that our proposed DEC achieves the highest Acc with low InvE. Furthermore, we notice that the \baselineB~is once again degrading the performance of the \generic~by distorting the input image. In~\tabref{tab:per_scale_imnet}, we report the performance of the baselines on Swin on different local scales. Our approach achieves the highest Acc on all local scales. Lastly, DEC produces the most consistent performance across different architectures, lowering the std of the Top-1 Acc to $1.49$ from $1.65$ of the \generic~model.

\begin{figure}[t]
    \centering
    \includegraphics[width=\linewidth]{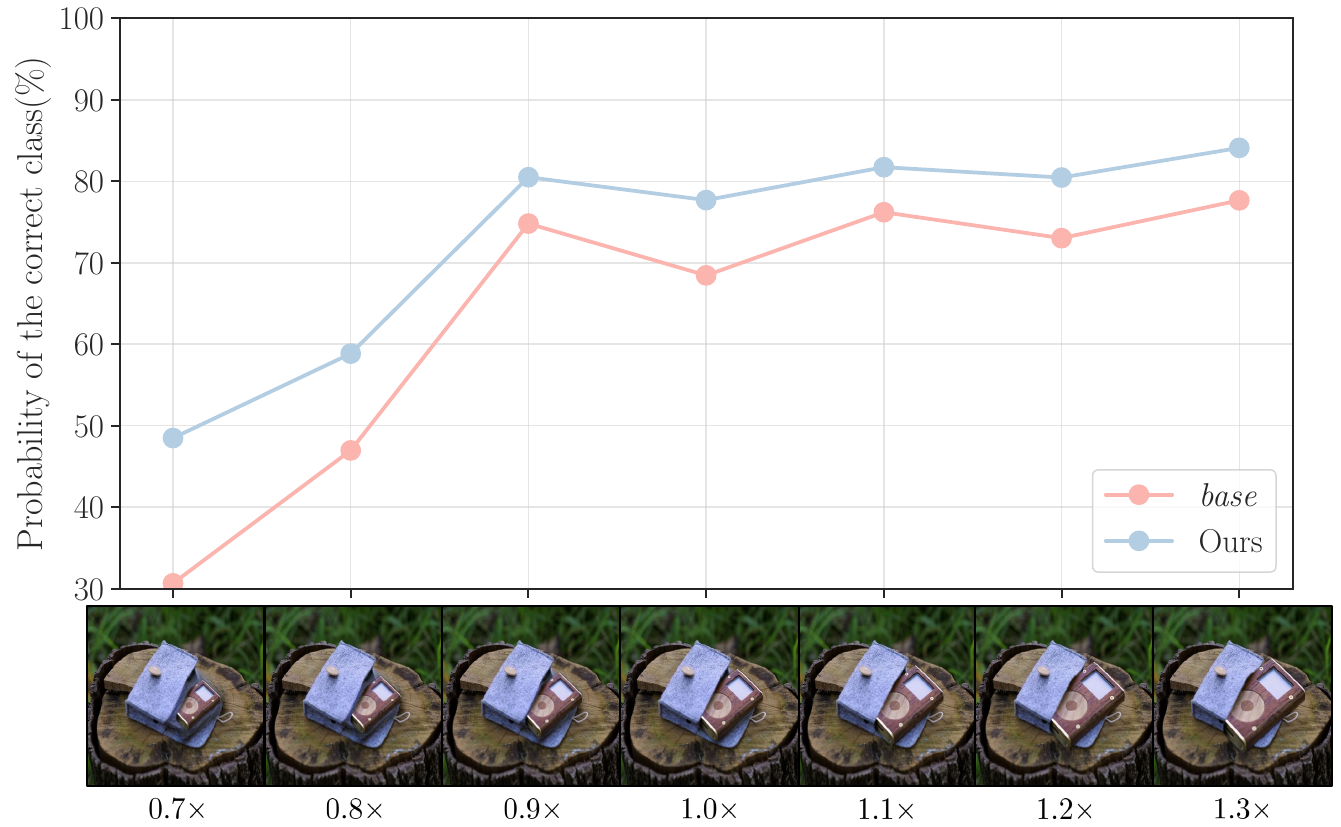}
    \vspace{-0.5cm}
    \caption{
        \textbf{Comparison on per-scale probability of correctness.} We locally scale the same input image within the range of $[0.7, 1.3]$ and report the probability of the correct class. Across all scales, our performance is $73.11 \pm 0.13$ while \generic's is $63.97 \pm 0.17$.
    }
    \vspace{-0.3cm}
    \label{fig:perscale_imnet_main}
\end{figure}

We visualize DEC's consistency across multiple scales in \figref{fig:perscale_imnet_main}. We show the probability of the correct class with the same image but locally scaled differently. We apply both methods to Swin~\cite{liu2021swin}. Compared to \generic, ours performs better on all scales and is more robust and invariant to challenging scenarios when the scale is extreme. In particular, when the input image is locally scaled by $0.7\times$, our method outperforms \generic~by 17.8$\%$.

\section{Conclusion}
In this work, we introduce the Deep Equilibrium Canonicalizer (DEC) to improve the local scale consistency of deep networks, tackling the fundamental challenge of scale variation in computer vision. We demonstrate that DEC can be adapted into existing deep-net architectures to improve both performance and scale consistency. Extensive experiments on three datasets and six architectures demonstrate its effectiveness. Notably, on the ImageNet dataset, DEC further improves the pre-trained model's accuracy, even for the original unscaled images. Finally, we demonstrate the potential of DEQ as an effective canonicalizer for achieving equivariance and hope that this work inspires future DEC efforts toward other equivariances.

\clearpage
{
    \small
    \bibliographystyle{ieeenat_fullname}
    \bibliography{main,scale_refs}

\begin{thebibliography}{76}
\providecommand{\natexlab}[1]{#1}
\providecommand{\url}[1]{\texttt{#1}}
\expandafter\ifx\csname urlstyle\endcsname\relax
  \providecommand{\doi}[1]{doi: #1}\else
  \providecommand{\doi}{doi: \begingroup \urlstyle{rm}\Url}\fi

\bibitem[Adelson et~al.(1984)Adelson, Anderson, Bergen, Burt, and Ogden]{adelson1984pyramid}
Edward~H Adelson, Charles~H Anderson, James~R Bergen, Peter~J Burt, and Joan~M Ogden.
\newblock Pyramid methods in image processing.
\newblock \emph{RCA engineer}, 1984.

\bibitem[Amos et~al.(2023)]{amos2023tutorial}
Brandon Amos et~al.
\newblock Tutorial on amortized optimization.
\newblock \emph{Foundations and Trends{\textregistered} in Machine Learning}, 2023.

\bibitem[Anderson(1965)]{anderson1965iterative}
Donald~G. Anderson.
\newblock Iterative procedures for nonlinear integral equations.
\newblock \emph{Journal of the ACM}, 1965.

\bibitem[Bai et~al.(2019)Bai, Kolter, and Koltun]{bai2019deep}
Shaojie Bai, J~Zico Kolter, and Vladlen Koltun.
\newblock Deep equilibrium models.
\newblock In \emph{NeurIPS}, 2019.

\bibitem[Bai et~al.(2020)Bai, Koltun, and Kolter]{bai2020multiscale}
Shaojie Bai, Vladlen Koltun, and J~Zico Kolter.
\newblock Multiscale deep equilibrium models.
\newblock In \emph{NeurIPS}, 2020.

\bibitem[Bai et~al.(2022)Bai, Geng, Savani, and Kolter]{bai2022deep}
Shaojie Bai, Zhengyang Geng, Yash Savani, and J~Zico Kolter.
\newblock Deep equilibrium optical flow estimation.
\newblock In \emph{CVPR}, 2022.

\bibitem[Bao et~al.(2022)Bao, Dong, and Wei]{bao2021beit}
Hangbo Bao, Li Dong, and Furu Wei.
\newblock {BEiT}: Bert pre-training of image transformers.
\newblock In \emph{ICLR}, 2022.

\bibitem[Barroso-Laguna et~al.(2022)Barroso-Laguna, Tian, and Mikolajczyk]{barroso2022scalenet}
Axel Barroso-Laguna, Yurun Tian, and Krystian Mikolajczyk.
\newblock Scalenet: A shallow architecture for scale estimation.
\newblock In \emph{Proceedings of the IEEE/CVF Conference on Computer Vision and Pattern Recognition}, pages 12808--12818, 2022.

\bibitem[Basu et~al.(2023)Basu, Sattigeri, Ramamurthy, Chenthamarakshan, Varshney, Varshney, and Das]{basu2023equi}
Sourya Basu, Prasanna Sattigeri, Karthikeyan~Natesan Ramamurthy, Vijil Chenthamarakshan, Kush~R Varshney, Lav~R Varshney, and Payel Das.
\newblock Equi-tuning: Group equivariant fine-tuning of pretrained models.
\newblock In \emph{AAAI}, 2023.

\bibitem[Basu et~al.(2024)Basu, Katdare, Sattigeri, Chenthamarakshan, Driggs-Campbell, Das, and Varshney]{basu2024efficient}
Sourya Basu, Pulkit Katdare, Prasanna Sattigeri, Vijil Chenthamarakshan, Katherine Driggs-Campbell, Payel Das, and Lav~R Varshney.
\newblock Efficient equivariant transfer learning from pretrained models.
\newblock In \emph{NeurIPS}, 2024.

\bibitem[Burt and Adelson(1983)]{1095851}
P. Burt and E. Adelson.
\newblock The {Laplacian} pyramid as a compact image code.
\newblock \emph{IEEE Transactions on Communications}, 1983.

\bibitem[Burt and Adelson(1987)]{burt1987laplacian}
Peter~J Burt and Edward~H Adelson.
\newblock The laplacian pyramid as a compact image code.
\newblock In \emph{Readings in computer vision}. 1987.

\bibitem[Cesa et~al.(2022)Cesa, Lang, and Weiler]{cesa2022program}
Gabriele Cesa, Leon Lang, and Maurice Weiler.
\newblock A program to build {E(N)}-equivariant steerable {CNNs}.
\newblock In \emph{ICLR}, 2022.

\bibitem[Chen et~al.(2017)Chen, Papandreou, Kokkinos, Murphy, and Yuille]{chen2017deeplab}
Liang-Chieh Chen, George Papandreou, Iasonas Kokkinos, Kevin Murphy, and Alan~L Yuille.
\newblock Deeplab: Semantic image segmentation with deep convolutional nets, atrous convolution, and fully connected crfs.
\newblock \emph{IEEE TPAMI}, 2017.

\bibitem[Cohen and Welling(2016)]{cohen2016group}
Taco Cohen and Max Welling.
\newblock Group equivariant convolutional networks.
\newblock In \emph{Proc. ICML}, 2016.

\bibitem[Dai et~al.(2017)Dai, Qi, Xiong, Li, Zhang, Hu, and Wei]{dai2017deformable}
Jifeng Dai, Haozhi Qi, Yuwen Xiong, Yi Li, Guodong Zhang, Han Hu, and Yichen Wei.
\newblock Deformable convolutional networks.
\newblock In \emph{ICCV}, 2017.

\bibitem[Deng et~al.(2009)Deng, Dong, Socher, Li, Li, and Fei-Fei]{deng2009imagenet}
Jia Deng, Wei Dong, Richard Socher, Li-Jia Li, Kai Li, and Li Fei-Fei.
\newblock {ImageNet}: A large-scale hierarchical image database.
\newblock In \emph{CVPR}, 2009.

\bibitem[Dosovitskiy et~al.(2021)Dosovitskiy, Beyer, Kolesnikov, Weissenborn, Zhai, Unterthiner, Dehghani, Minderer, Heigold, Gelly, et~al.]{dosovitskiy2020image}
Alexey Dosovitskiy, Lucas Beyer, Alexander Kolesnikov, Dirk Weissenborn, Xiaohua Zhai, Thomas Unterthiner, Mostafa Dehghani, Matthias Minderer, Georg Heigold, Sylvain Gelly, et~al.
\newblock An image is worth 16x16 words: Transformers for image recognition at scale.
\newblock In \emph{ICLR}, 2021.

\bibitem[Downs et~al.(2022)Downs, Francis, Koenig, Kinman, Hickman, Reymann, McHugh, and Vanhoucke]{downs2022google}
Laura Downs, Anthony Francis, Nate Koenig, Brandon Kinman, Ryan Hickman, Krista Reymann, Thomas~B McHugh, and Vincent Vanhoucke.
\newblock Google scanned objects: A high-quality dataset of 3d scanned household items.
\newblock In \emph{ICRA}, 2022.

\bibitem[Fan et~al.(2021)Fan, Xiong, Mangalam, Li, Yan, Malik, and Feichtenhofer]{fan2021multiscale}
Haoqi Fan, Bo Xiong, Karttikeya Mangalam, Yanghao Li, Zhicheng Yan, Jitendra Malik, and Christoph Feichtenhofer.
\newblock Multiscale vision transformers.
\newblock In \emph{Proc. ICCV}, 2021.

\bibitem[Grauman and Darrell(2005)]{grauman_2005_pyramid}
K. Grauman and T. Darrell.
\newblock The pyramid match kernel: discriminative classification with sets of image features.
\newblock In \emph{Proc. ICCV}, 2005.

\bibitem[Gu et~al.(2022)Gu, Kwon, Wang, Ye, Li, Chen, Lai, Chandra, and Pan]{gu2022multi}
Jiaqi Gu, Hyoukjun Kwon, Dilin Wang, Wei Ye, Meng Li, Yu-Hsin Chen, Liangzhen Lai, Vikas Chandra, and David~Z Pan.
\newblock Multi-scale high-resolution vision transformer for semantic segmentation.
\newblock In \emph{CVPR}, 2022.

\bibitem[Hartford et~al.(2018)Hartford, Graham, Leyton-Brown, and Ravanbakhsh]{hartford2018deep}
Jason Hartford, Devon Graham, Kevin Leyton-Brown, and Siamak Ravanbakhsh.
\newblock Deep models of interactions across sets.
\newblock In \emph{Proc. ICML}, 2018.

\bibitem[He et~al.(2015)He, Zhang, Ren, and Sun]{he2015spatial}
Kaiming He, Xiangyu Zhang, Shaoqing Ren, and Jian Sun.
\newblock Spatial pyramid pooling in deep convolutional networks for visual recognition.
\newblock \emph{IEEE TPAMI}, 2015.

\bibitem[He et~al.(2016)He, Zhang, Ren, and Sun]{he2016resnet}
Kaiming He, Xiangyu Zhang, Shaoqing Ren, and Jian Sun.
\newblock Deep residual learning for image recognition.
\newblock In \emph{CVPR}, 2016.

\bibitem[He et~al.(2022)He, Chen, Xie, Li, Doll{\'a}r, and Girshick]{he2022masked}
Kaiming He, Xinlei Chen, Saining Xie, Yanghao Li, Piotr Doll{\'a}r, and Ross Girshick.
\newblock Masked autoencoders are scalable vision learners.
\newblock In \emph{CVPR}, 2022.

\bibitem[Jaderberg et~al.(2015)Jaderberg, Simonyan, Zisserman, et~al.]{jaderberg2015spatial}
Max Jaderberg, Karen Simonyan, Andrew Zisserman, et~al.
\newblock Spatial transformer networks.
\newblock In \emph{NeurIPS}, 2015.

\bibitem[Kaba et~al.(2023)Kaba, Mondal, Zhang, Bengio, and Ravanbakhsh]{kaba2023equivariance}
S{\'e}kou-Oumar Kaba, Arnab~Kumar Mondal, Yan Zhang, Yoshua Bengio, and Siamak Ravanbakhsh.
\newblock Equivariance with learned canonicalization functions.
\newblock In \emph{ICML}, 2023.

\bibitem[Kanazawa et~al.(2014)Kanazawa, Sharma, and Jacobs]{kanazawa2014locally}
Angjoo Kanazawa, Abhishek Sharma, and David Jacobs.
\newblock Locally scale-invariant convolutional neural networks.
\newblock \emph{arXiv preprint arXiv:1412.5104}, 2014.

\bibitem[Kirillov et~al.(2023)Kirillov, Mintun, Ravi, Mao, Rolland, Gustafson, Xiao, Whitehead, Berg, Lo, et~al.]{kirillov2023segment}
Alexander Kirillov, Eric Mintun, Nikhila Ravi, Hanzi Mao, Chloe Rolland, Laura Gustafson, Tete Xiao, Spencer Whitehead, Alexander~C Berg, Wan-Yen Lo, et~al.
\newblock Segment anything.
\newblock In \emph{ICCV}, 2023.

\bibitem[Lazebnik et~al.(2006)Lazebnik, Schmid, and Ponce]{lazebnik2006beyond}
Svetlana Lazebnik, Cordelia Schmid, and Jean Ponce.
\newblock Beyond bags of features: Spatial pyramid matching for recognizing natural scene categories.
\newblock In \emph{Proc. CVPR}, 2006.

\bibitem[LeCun et~al.(1998)LeCun, Bottou, Bengio, and Haffner]{lecun1998mnist}
Yann LeCun, L{\'e}on Bottou, Yoshua Bengio, and Patrick Haffner.
\newblock Gradient-based learning applied to document recognition.
\newblock \emph{Proceedings of the IEEE}, 1998.

\bibitem[Lee et~al.(2022)Lee, Jeong, and Cho]{lee2022self}
Jongmin Lee, Yoonwoo Jeong, and Minsu Cho.
\newblock Self-supervised learning of image scale and orientation.
\newblock \emph{arXiv preprint arXiv:2206.07259}, 2022.

\bibitem[Lindeberg(1993)]{lindeberg2013scale}
Tony Lindeberg.
\newblock \emph{Scale-space theory in computer vision}.
\newblock 1993.

\bibitem[Liu et~al.(2020)Liu, Yeh, and Schwing]{liu2020pic}
Iou-Jen Liu, Raymond~A Yeh, and Alexander~G Schwing.
\newblock Pic: permutation invariant critic for multi-agent deep reinforcement learning.
\newblock In \emph{Proc. CORL}, 2020.

\bibitem[Liu et~al.(2021{\natexlab{a}})Liu, Ren, Yeh, and Schwing]{liu2021semantic}
Iou-Jen Liu, Zhongzheng Ren, Raymond~A Yeh, and Alexander~G Schwing.
\newblock Semantic tracklets: An object-centric representation for visual multi-agent reinforcement learning.
\newblock In \emph{Proc. IROS}, 2021{\natexlab{a}}.

\bibitem[Liu et~al.(2024)Liu, Zeng, Ren, Li, Zhang, Yang, Jiang, Li, Yang, Su, et~al.]{liu2024grounding}
Shilong Liu, Zhaoyang Zeng, Tianhe Ren, Feng Li, Hao Zhang, Jie Yang, Qing Jiang, Chunyuan Li, Jianwei Yang, Hang Su, et~al.
\newblock Grounding {DINO}: Marrying {DINO} with grounded pre-training for open-set object detection.
\newblock In \emph{ECCV}, 2024.

\bibitem[Liu et~al.(2021{\natexlab{b}})Liu, Lin, Cao, Hu, Wei, Zhang, Lin, and Guo]{liu2021swin}
Ze Liu, Yutong Lin, Yue Cao, Han Hu, Yixuan Wei, Zheng Zhang, Stephen Lin, and Baining Guo.
\newblock Swin transformer: Hierarchical vision transformer using shifted windows.
\newblock In \emph{ICCV}, 2021{\natexlab{b}}.

\bibitem[Lowe(1999)]{lowe1999object}
David~G Lowe.
\newblock Object recognition from local scale-invariant features.
\newblock In \emph{Proc. ICCV}, 1999.

\bibitem[Lowe(2004)]{lowe2004distinctive}
David~G Lowe.
\newblock Distinctive image features from scale-invariant keypoints.
\newblock \emph{IJCV}, 2004.

\bibitem[Ma et~al.(2024)Ma, Wang, Lim, Jegelka, and Wang]{ma2024canonicalization}
George Ma, Yifei Wang, Derek Lim, Stefanie Jegelka, and Yisen Wang.
\newblock A canonicalization perspective on invariant and equivariant learning.
\newblock In \emph{NeurIPS}, 2024.

\bibitem[Marr(1982)]{marr2010vision}
David Marr.
\newblock \emph{Vision: A computational investigation into the human representation and processing of visual information}.
\newblock 1982.

\bibitem[Micaelli et~al.(2023)Micaelli, Vahdat, Yin, Kautz, and Molchanov]{micaelli2023recurrence}
Paul Micaelli, Arash Vahdat, Hongxu Yin, Jan Kautz, and Pavlo Molchanov.
\newblock Recurrence without recurrence: Stable video landmark detection with deep equilibrium models.
\newblock In \emph{CVPR}, 2023.

\bibitem[Mondal et~al.(2023)Mondal, Panigrahi, Kaba, Mudumba, and Ravanbakhsh]{mondal2023equivariant}
Arnab~Kumar Mondal, Siba~Smarak Panigrahi, Oumar Kaba, Sai~Rajeswar Mudumba, and Siamak Ravanbakhsh.
\newblock Equivariant adaptation of large pretrained models.
\newblock In \emph{NeurIPS}, 2023.

\bibitem[Oquab et~al.(2024)Oquab, Darcet, Moutakanni, Vo, Szafraniec, Khalidov, Fernandez, Haziza, Massa, El-Nouby, et~al.]{oquab2023dinov2}
Maxime Oquab, Th{\'e}o Darcet, Theo Moutakanni, Huy Vo, Marc Szafraniec, Vasil Khalidov, Pierre Fernandez, Daniel Haziza, Francisco Massa, Alaaeldin El-Nouby, et~al.
\newblock {DINOv2}: Learning robust visual features without supervision.
\newblock \emph{TMLR}, 2024.

\bibitem[Panigrahi and Mondal(2024)]{panigrahi2024improved}
Siba~Smarak Panigrahi and Arnab~Kumar Mondal.
\newblock Improved canonicalization for model agnostic equivariance.
\newblock In \emph{CVPRW}, 2024.

\bibitem[Puny et~al.(2022)Puny, Atzmon, Ben-Hamu, Misra, Grover, Smith, and Lipman]{puny2021frame}
Omri Puny, Matan Atzmon, Heli Ben-Hamu, Ishan Misra, Aditya Grover, Edward~J Smith, and Yaron Lipman.
\newblock Frame averaging for invariant and equivariant network design.
\newblock In \emph{ICLR}, 2022.

\bibitem[Rahman and Yeh(2024)]{rahman2024truly}
Md~Ashiqur Rahman and Raymond~A Yeh.
\newblock Truly scale-equivariant deep nets with {Fourier} layers.
\newblock In \emph{Proc. NeurIPS}, 2024.

\bibitem[Rahman and Yeh(2025)]{rahman2025group}
Md~Ashiqur Rahman and Raymond~A. Yeh.
\newblock Group downsampling with equivariant anti-aliasing.
\newblock In \emph{The Thirteenth International Conference on Learning Representations}, 2025.

\bibitem[Ranftl et~al.(2021)Ranftl, Bochkovskiy, and Koltun]{ranftl2021vision}
Ren{\'e} Ranftl, Alexey Bochkovskiy, and Vladlen Koltun.
\newblock Vision transformers for dense prediction.
\newblock In \emph{ICCV}, 2021.

\bibitem[Ravanbakhsh et~al.(2017)Ravanbakhsh, Schneider, and Poczos]{ravanbakhsh_sets}
Siamak Ravanbakhsh, Jeff Schneider, and Barnabas Poczos.
\newblock Deep learning with sets and point clouds.
\newblock In \emph{Proc. ICLR workshop}, 2017.

\bibitem[Recasens et~al.(2018)Recasens, Kellnhofer, Stent, Matusik, and Torralba]{recasens2018learning}
Adria Recasens, Petr Kellnhofer, Simon Stent, Wojciech Matusik, and Antonio Torralba.
\newblock Learning to zoom: a saliency-based sampling layer for neural networks.
\newblock In \emph{ECCV}, 2018.

\bibitem[Rojas-Gomez et~al.(2022)Rojas-Gomez, Lim, Schwing, Do, and Yeh]{rojas2022learnable}
Renan~A Rojas-Gomez, Teck-Yian Lim, Alex Schwing, Minh Do, and Raymond~A Yeh.
\newblock Learnable polyphase sampling for shift invariant and equivariant convolutional networks.
\newblock In \emph{Proc. NeurIPS}, 2022.

\bibitem[Rojas-Gomez et~al.(2024)Rojas-Gomez, Lim, Do, and Yeh]{rojas2024making}
Renan~A Rojas-Gomez, Teck-Yian Lim, Minh~N Do, and Raymond~A Yeh.
\newblock Making vision transformers truly shift-equivariant.
\newblock In \emph{CVPR}, 2024.

\bibitem[Simoncelli and Freeman(1995)]{537667}
E.P. Simoncelli and W.T. Freeman.
\newblock The steerable pyramid: a flexible architecture for multi-scale derivative computation.
\newblock In \emph{Proc. ICIP}, 1995.

\bibitem[Sosnovik et~al.(2020)Sosnovik, Szmaja, and Smeulders]{sosnovik2019scale}
Ivan Sosnovik, Micha{\l} Szmaja, and Arnold Smeulders.
\newblock Scale-equivariant steerable networks.
\newblock In \emph{ICLR}, 2020.

\bibitem[Sosnovik et~al.(2021)Sosnovik, Moskalev, and Smeulders]{sosnovik2021disco}
Ivan Sosnovik, Artem Moskalev, and Arnold Smeulders.
\newblock {DISCO}: accurate discrete scale convolutions.
\newblock In \emph{Proc. BMVC}, 2021.

\bibitem[Suvorov et~al.(2022)Suvorov, Logacheva, Mashikhin, Remizova, Ashukha, Silvestrov, Kong, Goka, Park, and Lempitsky]{suvorov2022resolution}
Roman Suvorov, Elizaveta Logacheva, Anton Mashikhin, Anastasia Remizova, Arsenii Ashukha, Aleksei Silvestrov, Naejin Kong, Harshith Goka, Kiwoong Park, and Victor Lempitsky.
\newblock Resolution-robust large mask inpainting with {Fourier} convolutions.
\newblock In \emph{WACV}, 2022.

\bibitem[Thavamani et~al.(2021)Thavamani, Li, Cebron, and Ramanan]{thavamani2021fovea}
Chittesh Thavamani, Mengtian Li, Nicolas Cebron, and Deva Ramanan.
\newblock Fovea: Foveated image magnification for autonomous navigation.
\newblock In \emph{ICCV}, 2021.

\bibitem[Tian et~al.(2023)Tian, Wu, Dai, Hu, Qiao, and Jiang]{tian2023resformer}
Rui Tian, Zuxuan Wu, Qi Dai, Han Hu, Yu Qiao, and Yu-Gang Jiang.
\newblock Resformer: Scaling vits with multi-resolution training.
\newblock In \emph{CVPR}, 2023.

\bibitem[Touvron et~al.(2021)Touvron, Cord, Douze, Massa, Sablayrolles, and J{\'e}gou]{touvron2021deit}
Hugo Touvron, Matthieu Cord, Matthijs Douze, Francisco Massa, Alexandre Sablayrolles, and Herv{\'e} J{\'e}gou.
\newblock Training data-efficient image transformers \& distillation through attention.
\newblock In \emph{ICML}, 2021.

\bibitem[Van De~Ville et~al.(2008)Van De~Ville, Sage, Bala{\'c}, and Unser]{van2008marr}
Dimitri Van De~Ville, Daniel Sage, Katarina Bala{\'c}, and Michael Unser.
\newblock The {Marr} wavelet pyramid and multiscale directional image analysis.
\newblock In \emph{European Signal Processing Conference}, 2008.

\bibitem[Weiler et~al.(2018)Weiler, Hamprecht, and Storath]{weiler2018learning}
Maurice Weiler, Fred~A Hamprecht, and Martin Storath.
\newblock Learning steerable filters for rotation equivariant {CNNs}.
\newblock In \emph{Proc. CVPR}, 2018.

\bibitem[Wightman(2019)]{rw2019timm}
Ross Wightman.
\newblock Pytorch image models.
\newblock \url{https://github.com/huggingface/pytorch-image-models}, 2019.

\bibitem[Witkin et~al.(1987)Witkin, Terzopoulos, and Kass]{witkin1987signal}
Andrew Witkin, Demetri Terzopoulos, and Michael Kass.
\newblock Signal matching through scale space.
\newblock \emph{International Journal of Computer Vision}, 1987.

\bibitem[Witkin(1987)]{witkin1987scale}
Andrew~P Witkin.
\newblock Scale-space filtering.
\newblock In \emph{Readings in computer vision}. Elsevier, 1987.

\bibitem[Worrall and Welling(2019)]{worrall2019deep}
Daniel Worrall and Max Welling.
\newblock Deep scale-spaces: Equivariance over scale.
\newblock In \emph{Proc. NeurIPS}, 2019.

\bibitem[Xu et~al.(2021)Xu, Kim, Rainforth, and Teh]{xu2021group}
Jin Xu, Hyunjik Kim, Thomas Rainforth, and Yee Teh.
\newblock Group equivariant subsampling.
\newblock In \emph{Proc. NeurIPS}, 2021.

\bibitem[Xu et~al.(2014)Xu, Xiao, Zhang, Yang, and Zhang]{xu2014scale}
Yichong Xu, Tianjun Xiao, Jiaxing Zhang, Kuiyuan Yang, and Zheng Zhang.
\newblock Scale-invariant convolutional neural networks.
\newblock \emph{arXiv preprint arXiv:1411.6369}, 2014.

\bibitem[Yeh et~al.(2019{\natexlab{a}})Yeh, Hu, and Schwing]{yeh2019chirality}
Raymond~A Yeh, Yuan-Ting Hu, and Alexander Schwing.
\newblock Chirality nets for human pose regression.
\newblock In \emph{Proc. NeurIPS}, 2019{\natexlab{a}}.

\bibitem[Yeh et~al.(2019{\natexlab{b}})Yeh, Schwing, Huang, and Murphy]{yeh2019diverse}
Raymond~A Yeh, Alexander~G Schwing, Jonathan Huang, and Kevin Murphy.
\newblock Diverse generation for multi-agent sports games.
\newblock In \emph{Proc. CVPR}, 2019{\natexlab{b}}.

\bibitem[Yeh et~al.(2022)Yeh, Hu, Hasegawa-Johnson, and Schwing]{yeh2022equivariance}
Raymond~A Yeh, Yuan-Ting Hu, Mark Hasegawa-Johnson, and Alexander Schwing.
\newblock Equivariance discovery by learned parameter-sharing.
\newblock In \emph{Proc. AISTATS}, 2022.

\bibitem[Zaal(2020)]{zaal2020hdri}
Greg Zaal.
\newblock Hdri haven: Free high-quality hdr images under a cc0 license, 2020.
\newblock Accessed via Poly Haven (formerly HDRI Haven).

\bibitem[Zaheer et~al.(2017)Zaheer, Kottur, Ravanbakhsh, Poczos, Salakhutdinov, and Smola]{zaheer2017deep}
Manzil Zaheer, Satwik Kottur, Siamak Ravanbakhsh, Barnabas Poczos, Ruslan~R Salakhutdinov, and Alexander~J Smola.
\newblock Deep sets.
\newblock In \emph{Proc. NeurIPS}, 2017.

\bibitem[Zhang(2019)]{zhang2019making}
Richard Zhang.
\newblock Making convolutional networks shift-invariant again.
\newblock In \emph{Proc. ICML}, 2019.

\bibitem[Zhao et~al.(2017)Zhao, Shi, Qi, Wang, and Jia]{zhao2017pyramid}
Hengshuang Zhao, Jianping Shi, Xiaojuan Qi, Xiaogang Wang, and Jiaya Jia.
\newblock Pyramid scene parsing network.
\newblock In \emph{Proc. CVPR}, 2017.

\end{thebibliography}
}
\clearpage
\appendix
\clearpage
\section*{Appendix}

\setcounter{section}{0}
\renewcommand{\theHsection}{A\arabic{section}}
\renewcommand{\thesection}{A\arabic{section}}
\renewcommand{\thetable}{A\arabic{table}}
\setcounter{table}{0}
\setcounter{figure}{0}
\renewcommand{\thetable}{A\arabic{table}}
\renewcommand\thefigure{A\arabic{figure}}
\renewcommand{\theHtable}{A.Tab.\arabic{table}}%
\renewcommand{\theHfigure}{A.Abb.\arabic{figure}}%
\renewcommand\theequation{A\arabic{equation}}
\renewcommand{\theHequation}{A.Abb.\arabic{equation}}%

The appendix is organized as follows:
\begin{itemize}
    \item In~\secref{app:supp_impl}, we provide additional implementation details of our method.
    \item In ~\secref{app:group}, we provide additional details of our monotone scaling group and its parameterization.
    \item In~\secref{app:proof}, we provide the detailed proofs for our lemmas in~\secref{app:runtime}, we discuss the runtime of the DEC module.
    \item In~\secref{app:results}, we provide additional analysis and results.
\end{itemize}

\section{Implementation details}\label{app:supp_impl}
\myparagraph{DEC module} For all experiments, we model the DEC module as $2$ layered CNN mapping with $64$ and $128$ channels. We perform adaptive pooling at the end to get the desired number of monotone scaling parameters.

\myparagraph{Vanilla Canonicalization} To train vanilla canonicalization, we discretized the monotone scale parameter into $64$ different configurations for the locally scaled MNIST and object segmentation and $25$ different configurations for locally scaled ImageNet. The learnable energy functions are implemented via a $3$ layer CNN.

\myparagraph{Locally Scaled Object Segmentation} The pretrained models ViT~\cite{dosovitskiy2020image}, Swin~\cite{liu2021swin},  and DINOv2~\cite{oquab2023dinov2} are finetuned on the training set for $120$ epochs. We set the initial learning rate at $1\mathrm{e}{-4}$ and scaled it by a factor of $0.7$ for each $30$ epoch. We use DPT~\cite{ranftl2021vision} style segmentation head. We use the per-pixel cross-entropy loss to train each model. We set the weight for the ``background'' pixel class to $0.1$, and the weights of all other object classes are set to $1$. All baselines are finetuned for $60$ epochs with an initial learning rate of $2\mathrm{e}{-5}$

\myparagraph{Locally Scaled MNIST} Each architecture is trained for 50 epochs. We set the initial learning rate at $1\mathrm{e}{-4}$ for ResNet~\cite{he2016resnet}, ViT~\cite{dosovitskiy2020image}, DeiT~\cite{touvron2021deit}, and BEiT~\cite{bao2021beit}; $2\mathrm{e}{-5}$ for DINOv2~\cite{oquab2023dinov2}; and $8\mathrm{e}{-4}$ for Swin~\cite{liu2021swin}. All baselines are fine-tuned for $40$ epochs.

\myparagraph{Locally Scaled ImageNet} All baselines are fine-tuned following standard data augmentation practice for ImageNet finetuning \cite{he2022masked, liu2021swin}. We use a batch size of $80$ and an initial learning rate of $1\mathrm{e}{-7}$ with $5$ warm-up epochs. We list common training hyperparameters in~\cref{table:hyper_imnet}.
\begin{table}[h!]
\centering
\begin{tabular}{lc}
\toprule
\textbf{Hyperparameter} & \textbf{Value} \\
\midrule
Scheduler  & Cosine \\
Weight decay  & 0.05 \\
Warmup epochs  & 5 \\
Mixup  Alpha & 0.8 \\
Label smoothing  & 0.1 \\
Random Erase Prob  & 0.25 \\
Layer Decay Factor & 0.75\\
Epochs & 20\\
\bottomrule
\end{tabular}
\caption{ImageNet Finetuning Parameters}
\label{table:hyper_imnet}
\end{table}

\section{Monotone scaling group}
\label{app:group}

\subsection{Group axioms}
Our monotone scaling set $L$ must meet the following axioms to be considered as a group.
Here, ``$\cdot$'' represents the group product described in \equref{eqn:group_composition}.
The 4 axioms are as follows:
\begin{itemize}[leftmargin=*,topsep=-0.5pt]
    \item \textbf{Closure:} $\forall a, b \in L, a\cdot b \in L$
    \item \textbf{Associativity:} $\forall a, b, c \in L, a \cdot (b\cdot c) = (a \cdot b) \cdot c$
    \item \textbf{Existence of Identity:} $\exists e \in L : \forall a \in L,~~ e \cdot a = a$
    \item \textbf{Existence of Inverse:} $\forall a \in L, \exists a^{-1} \in L: a \cdot a^{-1} = e$.
\end{itemize}

\subsection{2D Monotone scaling group} 
\label{app:2d_scal_group}

\myparagraph{Construction of monotone scaling group in 2D} To form the monotone scaling group on $2D$, the set $$L_{\text{2D}}: \{l:[0,1]^2 \rightarrow [0,1]^2\}$$ should satisfy the two following properties:
\begin{itemize}
    \item For all $x \in [0,1]^2$ there is a neighborhood $\gW \subset [0,1]^2$ such that any function $l \in L_{\text{2D}}$ can be approximated by a linear function with Jacobian $J_l(x) \in \text{SPD}(2)$, \ie,  $2 \times 2$ symmetric positive definite Jacobian. This condition imposes local monotonicity on $l$.
    \item For all $l_1, l_2 \in L_{\text{2D}}$, their local Jacobian $J_{l_1}(x_1)$ and $J_{l_2}(x_2)$ commutes for $x_1, x_2 \in [0,1]^2$.
\end{itemize}
Formally, we state this in Lemma~\ref{lemma:group}.
\begin{mylemmae}
\begin{restatable}[]{mylemma}{lemmaone}\label{lemma:group}
The set of all locally monotone increasing functions $L_{\text{2D}}$ with commutative Jacobian is a valid group under the binary operation of function composition.
\end{restatable}
\end{mylemmae}
\begin{proof}
We provide the detailed proof in \secref{app;group_proof}.
\end{proof}

{\bf \noindent Parametrization of $l$.}
We parameterize $l$ through a set of independent monotone functions along the $x$ and $y$ axes via bilinear interpolation. Specifically, we decompose the function $l$ into two functions as 
\bea
l(x,y) = (l_X(x,y), l_Y(x,y)),
\eea
where $l_{X}, l_{Y}: [0,1]^2 \rightarrow [0,1]$. We parameterize each of them as piecewise linear functions. To achieve this we discretize the domain $[0,1]^2$ into uniform grid $\gG = \gX \times \gY$ where $\gX=\{x_0=0, x_1, \dots, x_N=1\}$ and $\gY=\{y_0=0, y_1, \dots, y_M=1\}$. We assume the set is ordered,~\ie, $x_i < x_j$ when $i<j$. 

We define independent monotone functions along each row $y_j$ and each column $x_i$ of the grid $\gG$ as follows:

For $x \in [x_{n-1}, x_n)$, the monotone function along row  $y_j$ is given by:  
\bea
l^{y_j}(x) =  \phi^{y_j}_{x_{n-1}} + \frac{\phi^{y_j}_{x_n} - \phi^{y_j}_{x_{n-1}}}{x_n - x_{n-1}}  \times (x - x_{n-1}).
\eea

Similarly, for $y \in [y_{m-1}, y_m)$, the monotone function along column $x_i$ is given by: \bea
l^{x_i}(y) =  \phi^{x_i}_{y_{m-1}} + \frac{\phi^{x_i}_{y_m} - \phi^{x_i}_{y_{m-1}}}{y_m - y_{m-1}}  \times (y - y_{m-1})
\eea

Here, $\phi$s are the learnable parameters and to preserve monotonicity we impose restriction  $\phi_{x_{n-1}}^{y_j} \le \phi_{x_n}^{y_j}~ \forall j, n$ and  $\phi_{y_{m-1}}^{x_i} \le \phi_{y_m}^{x_i}~ \forall i, m$.

Finally, we obtain $l_X(x,y)$ from $l^{y_i}s$ via linear interpolation as
\bea
l_X(x, y) = l^{y_{j-1}}(x) + \frac{l^{y_j}(x) - l^{y_{j-1}}(x)}{y - y_{j-1}} 
\eea 
when $y \in [y_{j-1} \le y_{j})$. We defined $l_Y(x,y)$ similarly. We approximate the inverse function $l^{-1}$ by computing the inverses of each $l^{x_i}$ and $l^{y_j}$ individually.

\section{Complete Proofs of Lemmas and Claims}
\label{app:proof}

\subsection{Proof of Lemma \ref{lemma:group_1d}}
\label{app:group_proof_id}
\begin{mylemmae}
\lemmaoned*
\end{mylemmae}
\begin{proof}

To prove that the set of all continuous strictly monotonic increasing functions $ L $ forms a group under function composition, we need to verify four properties: closure, associativity, identity element, and inverse element.

\myparagraph{Closure}
Let $ l_1, l_2 \in L $. Since $ l_1 $ and $ l_2 $ are continuous and strictly increasing, for any $ x_1, x_2 \in [0,1]$ if $ x_1 < x_2 $, then $l_1(x_1) < l_1(x_2)$. Thus, $ l_2(L_1(x_1)) < l_2(l_1(x_2)) $, showing that $ l_1 \circ l_2 $ is strictly increasing. Moreover, $ l_1 \circ l_2 $ is continuous because both are continuous.

\myparagraph{Associativity}
Function composition is inherently associative, thus satisfying the property.

\myparagraph{Identity Element}
The identity function is continuous and strictly increasing, so also an element of $L$

\myparagraph{Inverse Element}
Strictly monotone functions have an inverse, and the inverse is also monotonic. Thus, the inverse is also an element of $L$.

Therefore, we conclude that $L$ is a group.
\end{proof}

\subsection{Proof of Lemma \ref{lemma:group}}
\label{app;group_proof}
\begin{mylemmae}
\lemmaone*
\end{mylemmae}
\begin{proof}
To verify that $L_{\text{2D}}$ forms a group, we check the following properties:  

\parag{Closure:} For any $l_1, l_2 \in L_{\text{2D}}$, their composition $l_1 \circ l_2$ is also a locally monotone-invertible function.

Because the local Jacobian of the composition  $l_1 \circ l_2$ is 
\bea
J_{l_1 \circ l_2}(x) = J_{l_1}(l_2(x)) \cdot J_{l_2}(x) \quad \forall x \in [0,1]^2.
\eea
Since $J_{l_1}, J_{l_2} \in \text{SPD}(2)$ and commutes, their product $J_{l_1} \cdot J_{l_2}$ is also a SPD matrix. 
The composition of invertible functions is also invertible. And the $J_{l_1 \circ l_2}$ commutes due to associativity of matrix product.
Thus $l_1 \circ l_2 \in L_{\text{2D}}$.

\parag{Associativity:} Function composition is inherently associative, \ie, for all  $l_1, l_2, l_3 \in L_{\text{2D}}$, we have  
\bea
&(l_1 \circ l_2) \circ l_3 = l_1 \circ (l_2 \circ l_3).
\eea  
\parag{Existence of Identity:} The identity function $l$ has $2 \times 2$ identity matrix as local Jacobin. Thus, it maintains all the conditions of $L_{\text{2D}}$.

\parag{Existence of Inverse:} The inverse function of any $l \in L_{\text{2D}}$ can be obtained by inverting the local Jacobians. Specifically, for any $l \in L_{\text{2D}}$, the inverse $l^{-1}$ exists and satisfies:
\bea
J_{l^{-1}}(\mathbf{x}) = J_l(l^{-1}(\mathbf{x}))^{-1} \in \text{SPD}(2) \quad \forall x \in [0,1]^2
\eea
Furthermore, $J_{l^{-1}}J_{l_k} = J_{l_k}J_{l^{-1}}$ for any $l_k \in L_{\text{2D}}$ as
\begin{flalign}
& J_l J_{l_k} = J_{l_k} J_l \\
&\Rightarrow J_l^{-1} (J_l J_{l_k}) = J_l^{-1} (J_{l_k} J_l), \text{(left multiply by } J_l^{-1}\text{)}\\
&\Rightarrow J_{l_k} = J_l^{-1} J_{l_k} J_l \\
&\Rightarrow J_{l_k} J_l^{-1} = J_l^{-1} J_{l_k} \text{(right multiply by } J_l^{-1}\text{)}.
\end{flalign}

Thus, $l^{-1} \in L_{\text{2D}}$.

Therefore, $L_{\text{2D}}$ satisfies all group axioms, completing the proof.
\end{proof}

\section{Runtime of the DEC Module}
\label{app:runtime}
We use Anderson Acceleration to approximate the fixed point of the DEC. This requires a fixed number of forward passes through the lightweight DEC module. The computational complexity of this iterative process is $\mO(j T_{\text{DEC}})$, where $j$ is a fixed number of required iterations and $T_{\text{DEC}}$ is the computation cost associated with a single forward pass of the DEC module. The hyperparameter $j$ governs the trade-off between computational cost and the accuracy of the fixed-point approximation.

Empirically, for DINO-v2, the DEC module requires $24\%$ ($0.16$ sec) of the total required time ($0.66$ sec) to process a batch of $128$ images of size $224\times224$.

\section{Additional Results}
\label{app:results}
\subsection{Additional Baselines}
\label{sec:additional_baselines}
To evaluate the effectiveness of the DEC module in models with handcrafted hierarchical feature processing or image pyramid structures, we adapt our approach to HRViT \cite{gu2022multi} and ResFormer \cite{tian2023resformer} and report the results in \tabref{tab:pyramid}.

\begin{table}[t]
\centering
\small
\resizebox{0.8\linewidth}{!}{
    \begin{tabular}{lcc}
    \toprule
    Method & HRVit \cite{gu2022multi} & Resformer \cite{tian2023resformer} \\
    \midrule
    \baselineA  & 93.22 & 91.04\\
    \midrule
    \baselineB  & 94.93  & 95.27\\
    \baselineC  & 95.91  & 94.92\\
    \rowcolor{OursColor} Ours    & \bf 96.67 & \bf 96.91\\
    \bottomrule
    \end{tabular}
}
\vspace{-0.2cm}
\captionof{table}{Hierarchical baselines on scale-MNIST}
\label{tab:pyramid}
\end{table}

\subsection{Additional Ablation Study}
\label{sec:additional_ablation}
We perform additional ablation studies on the scale-MNIST dataset to evaluate the effect of (i) the number of DEC modules and (ii) the number of layers within each DEC module. The results are summarized in \tabref{tab:ablation_1}. We observe that increasing the number of DEC modules, \ie, repeatedly canonicalizing features throughout the network, improves performance compared to applying canonicalization only at the input level. 
\begin{table}[t]
\centering
\small
\begin{tabular}{cccccc}
\toprule
& & \multicolumn{4}{c}{\# layers in DEC Mod.}
\\
 & & 1 & 2 & 3 & 4 \\
\midrule
\multirow{4}{*}{\rotatebox[origin=c]{90}{\# DEC Mod}}
& 1 & 93.59  & 94.47  & 94.04 &  93.22  \\
& 2 & 94.98  & 96.04  & 95.16   &  95.25 \\
& 3 & 96.66  & 96.17  & 96.67  &  96.38  \\
& 4 & 96.58  & 96.78  & 96.65  &  96.50  \\
\bottomrule
\end{tabular}
\vspace{-0.3cm}
\caption{Ablation on the number of DEC modules and layers per module on scale-MNIST}
\label{tab:ablation_1}
\end{table}

We provide an additional ablation study on the choice of grid size for local scaling and report the results in \tabref{tab:ablation_2}. We observe that increasing the grid size improves the performance as it allows more flexible spatial parameterization of the local scaling operations.
\begin{table}[t]
\centering
\small

    \begin{tabular}{c ccccc}
    \toprule
    & & \multicolumn{4}{c}{Multiples ($\times$) of grid} \\
    & & 1.0 & 1.5 & 2.0 & 2.5 
    \\
    \midrule
    \multirow{2}{*}{\rotatebox[origin=c]{90}{layers}}
    & 2 & 96.61 & 96.95 & 97.06 & 97.94
    \\
    & 3 & 96.62 & 96.91 & 97.08 & 97.08
    \\
    \bottomrule
    \end{tabular}
\vspace{-0.2cm}
\caption{Acc. at multiple of initial grid size of $4$ with varying number of layers in DEC on scale-MNIST. }
\label{tab:ablation_2}
\end{table}

To assess potential side effects of scale equivariance, we report the accuracies of the adapted models on images of scale $1$, \ie, unmodified images in \tabref{tab:scale_1_ImNet}. We do not observe any drop in the performance.
\begin{table}[t]
\centering
\small
    \begin{tabular}{lcccc}
    \toprule
    Methods & ViT   & DeiT  & Swin  & BEiT  \\
    \midrule
    \generic    & 81.29 & 70.67 & 79.55 & 85.79 \\
    \baselineA     & 81.29 & 70.70 & 79.56 & 85.66 \\
    \midrule
    \baselineB   & 79.23 & 66.92 & 76.04 & 84.29 \\
    \baselineC    & 81.29 & 70.71 & 79.58 & 85.66 \\
    \rowcolor{OursColor} Ours    & \bf 81.43 & \bf 70.92 & \bf 79.86 & \bf 86.04\\
    \bottomrule
    \end{tabular}
\vspace{-0.3cm}
\captionof{table}{Acc. on unmodified (scale-$1$) ImageNet images.}
\label{tab:scale_1_ImNet}
\end{table}

\subsection{Additional Visualizations}
\begin{figure*}[h!]
    \centering
    \setlength{\tabcolsep}{3pt}
    \begin{tabular}{cc}
        \includegraphics[width=0.48\linewidth]{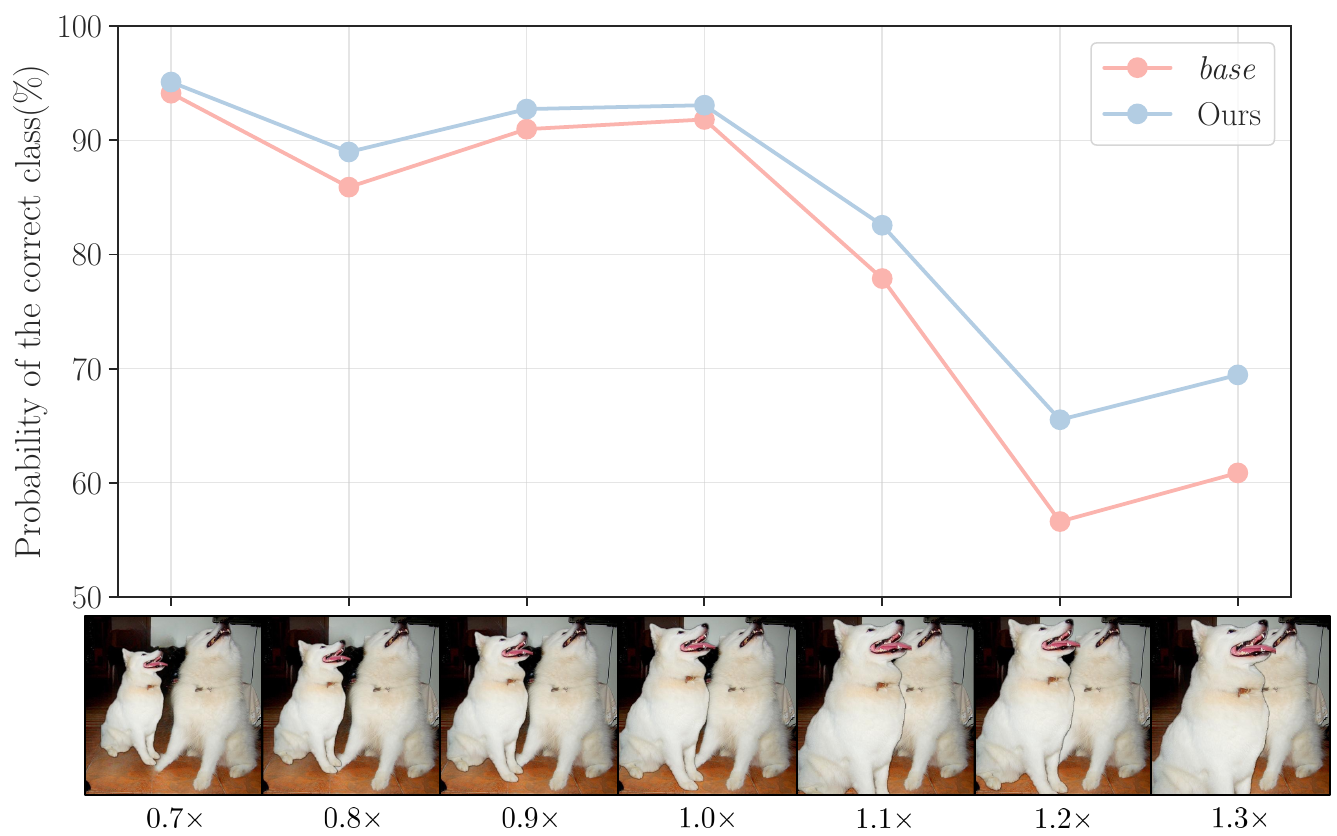}
        &
        \includegraphics[width=0.48\linewidth]{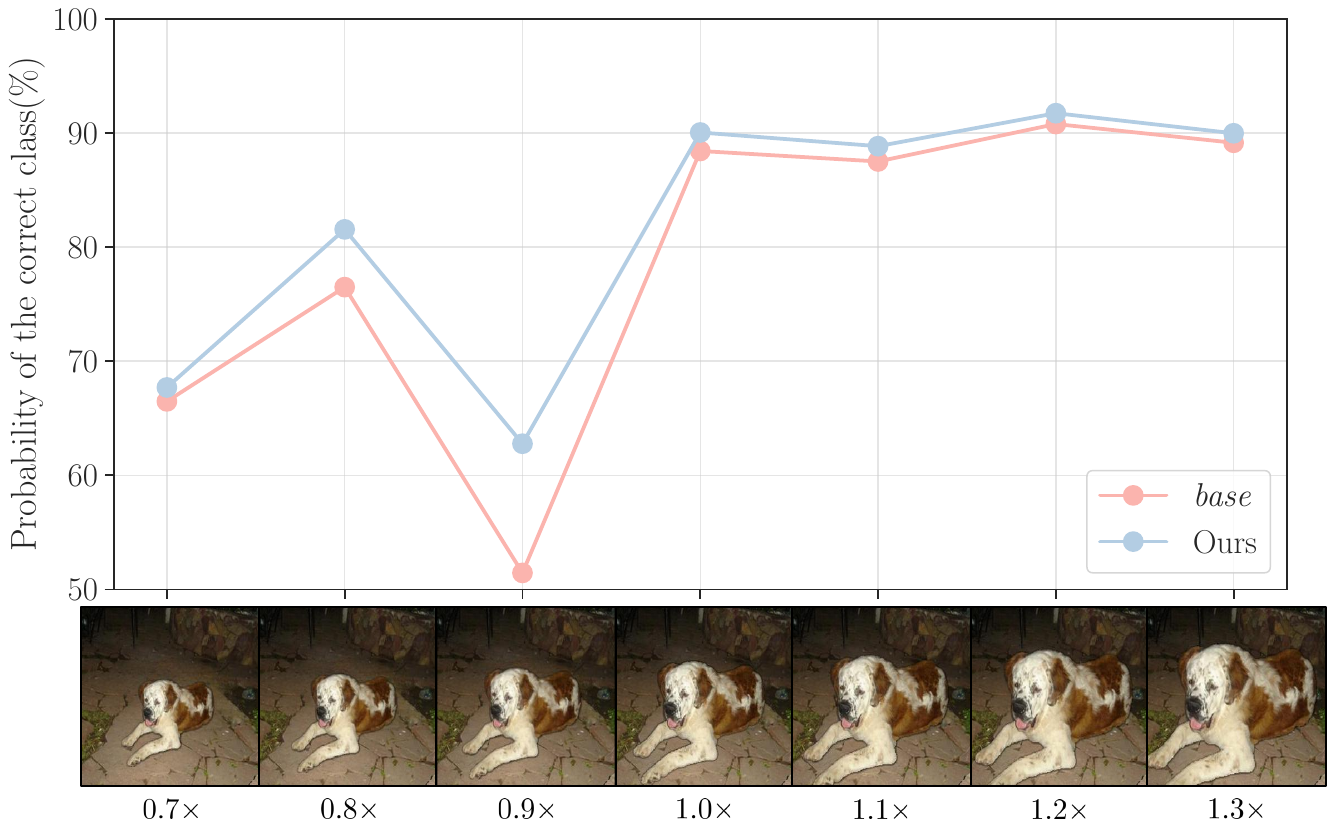}
        \\
        \includegraphics[width=0.48\linewidth]{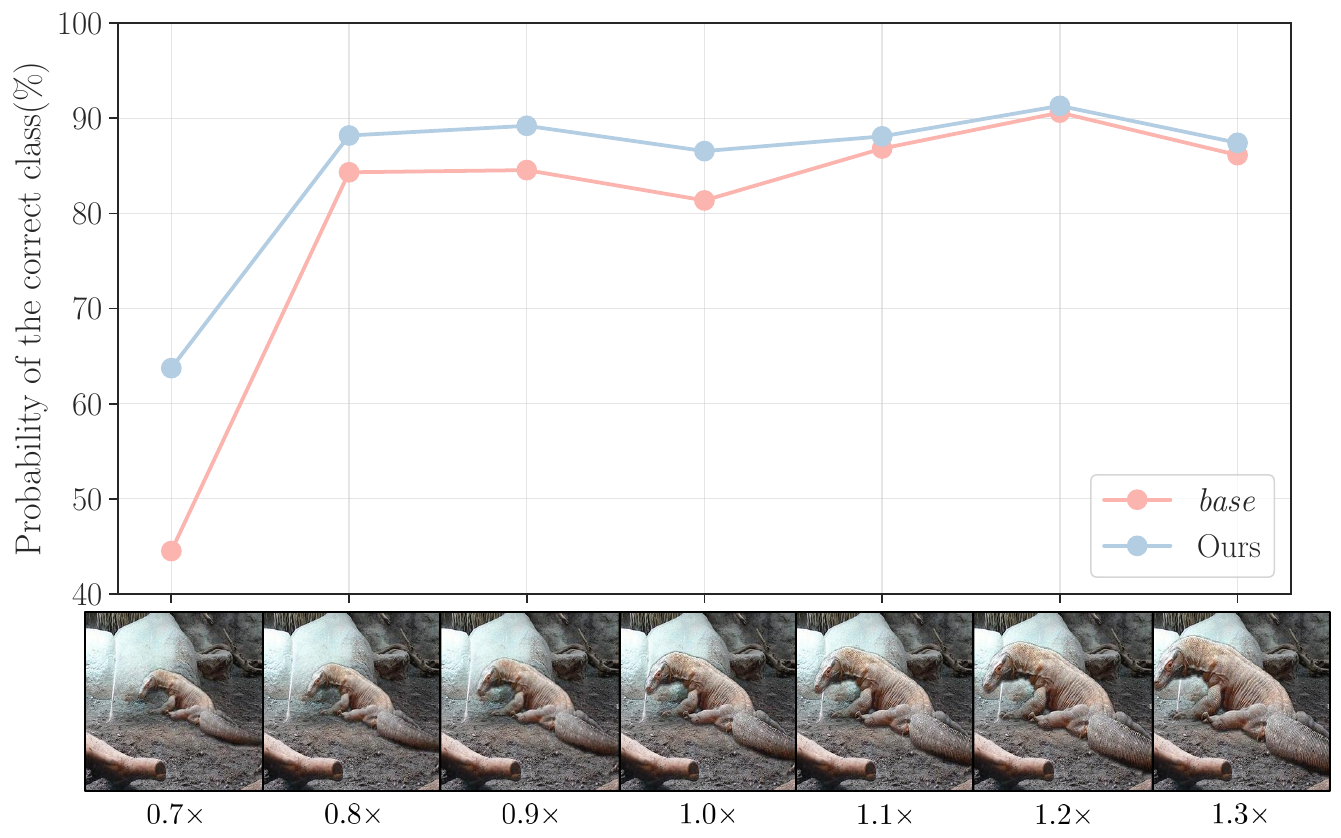}
        &
        \includegraphics[width=0.48\linewidth]{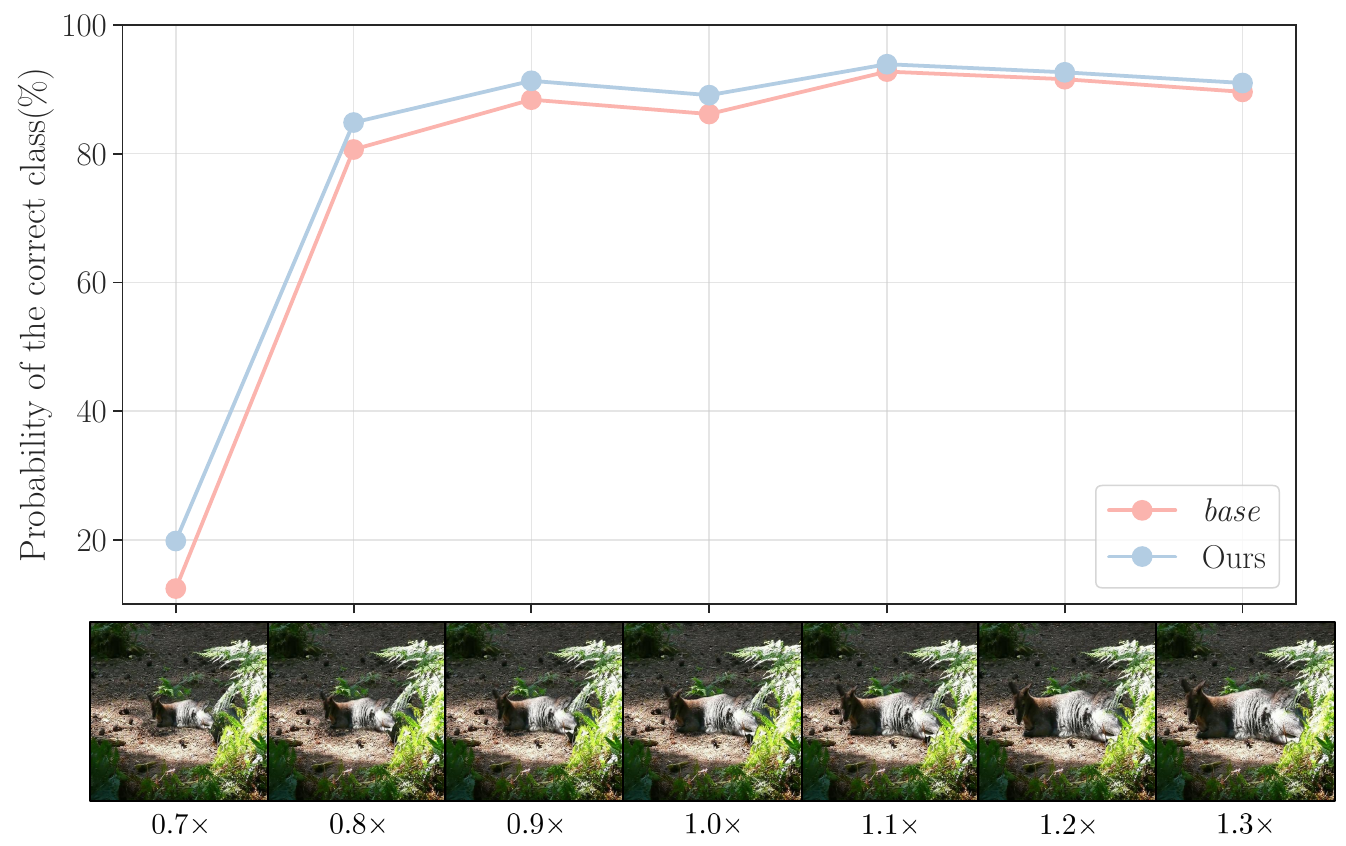}
        \\
    \end{tabular}
    \caption{
        \textbf{Comparison on per-scale probability of correctness.} We locally scale the same input image within the range of $[0.7, 1.3]$ and report the probability of the correct class.
    }
    \label{fig:perscale_imnet_supp}
\end{figure*}

Following the settings of~\figref{fig:perscale_imnet_main}, we report the results for more input images in~\figref{fig:perscale_imnet_supp}.
We observe that Ours is consistently better and more robust on all scales in comparison to \generic; especially on the more extreme local scale factors.

We present visualizations of learned monotone scaling by the DEC trained on MNIST in~\figref{fig:learned_warp}. We observe that the DEC module has learned to stretch/squeeze regions of the digits. However, the exact reasoning on why such scaling is beneficial to the deep-net remains challenging. The interpretability of the choice of learned canonical elements in a group is largely underexplored in the literature.

\begin{table*}[t]
    \centering
    \renewcommand{\arraystretch}{1}
    \begin{tabular}{c@{\hskip 1pt}c@{\hskip 1pt}c@{\hskip 1pt}}
        \includegraphics[width=0.32\linewidth, height=0.23\linewidth]{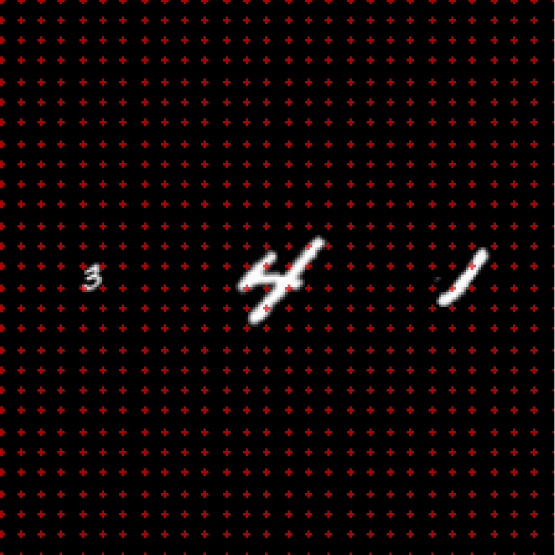} &
        \includegraphics[width=0.32\linewidth, height=0.23\linewidth]{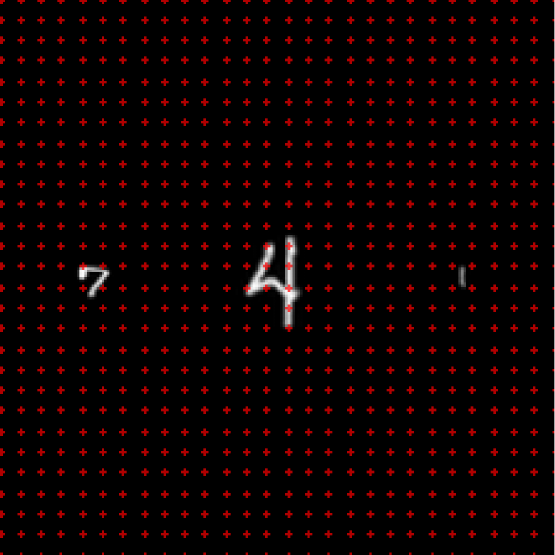} &
        \includegraphics[width=0.32\linewidth, height=0.23\linewidth]{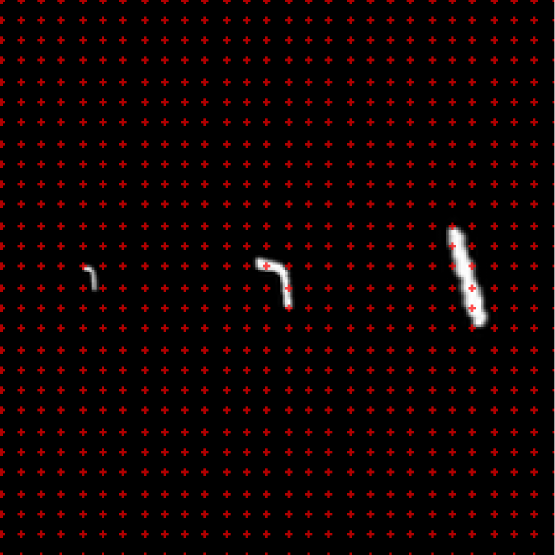}\\
        \multicolumn{3}{c}{Original Images}\\
        \includegraphics[width=0.32\linewidth, height=0.23\linewidth]{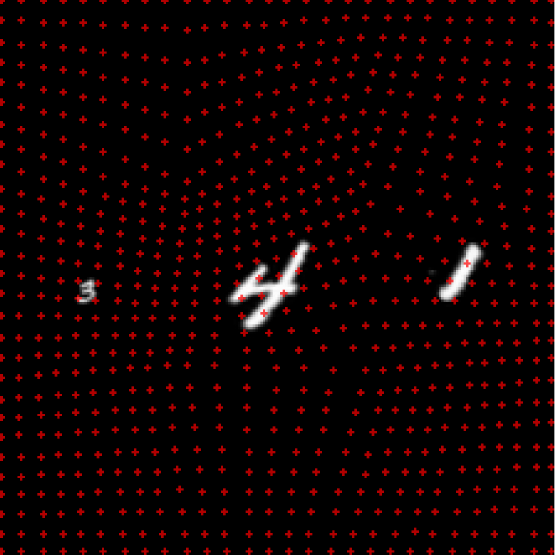} &
        \includegraphics[ width=0.32\linewidth, height=0.23\linewidth]{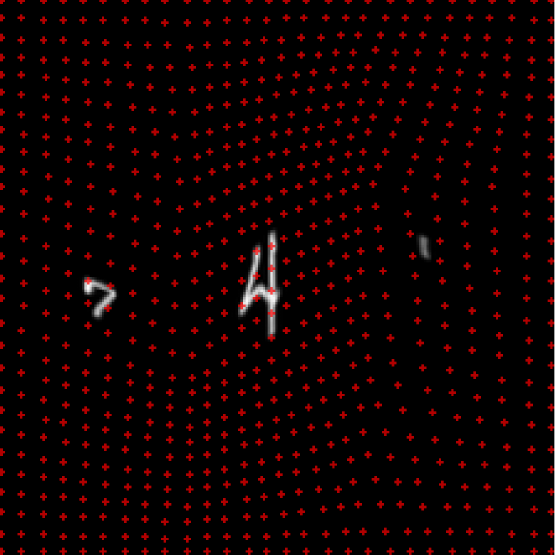} &
        \includegraphics[width=0.32\linewidth, height=0.23\linewidth]{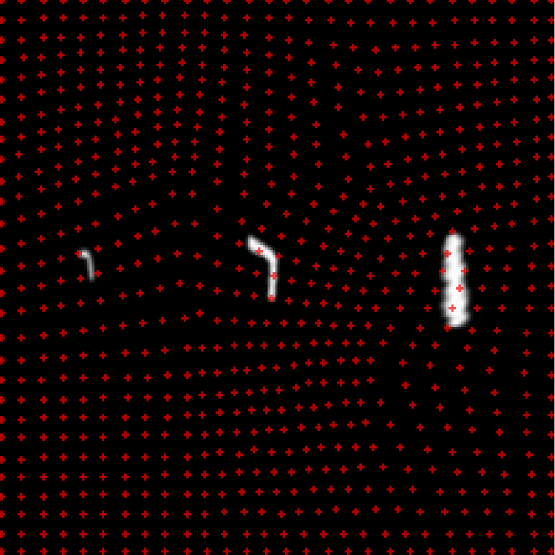} \\
        \multicolumn{3}{c}{Monotone Scaled Images}\\
    \end{tabular}
    \captionof{figure}{
         \textbf{Learned monotone scaling on locally scaled MNIST.} 
        We observe that stretching/squeezing is performed on the area with digits.
    }
    \label{fig:learned_warp}
\end{table*}

\end{document}